\documentclass[10pt,twocolumn,letterpaper]{article}

\usepackage[pagenumbers]{cvpr}

\usepackage{lineno}
\usepackage{multirow}
\usepackage{pifont}
\usepackage{graphicx}
\usepackage[dvipsnames]{xcolor}
\usepackage{amsmath}
\usepackage{amssymb}
\usepackage{booktabs}
\usepackage{overpic}
\usepackage{float}
\usepackage[linesnumbered, boxed, ruled]{algorithm2e}
\usepackage{array}
\usepackage{soul}
\usepackage{makecell}
\usepackage[table]{xcolor}
\usepackage{tcolorbox}
\usepackage{balance}
\usepackage[normalem]{ulem}
\useunder{\uline}{\ul}{}
\usepackage{leftindex}
\usepackage{arydshln}
\usepackage{enumitem}
\usepackage{siunitx}
\usepackage{makecell}
\usepackage[accsupp]{axessibility}  

\usepackage{bibunits}
\defaultbibliographystyle{ieeenat_fullname}
\defaultbibliography{main}

\definecolor{boost}{RGB}{243,220,205}
\definecolor{adapt}{RGB}{232,243,225}
\definecolor{boost_line}{RGB}{197,90,17}
\definecolor{adapt_line}{RGB}{84,130,53} 

\definecolor{cvprblue}{rgb}{0.21,0.49,0.74}
\usepackage[pagebackref,breaklinks,colorlinks,allcolors=cvprblue]{hyperref}

\def\paperID{} 
\def\confName{CVPR}
\def\confYear{2026}

\title{Restore Text First, Enhance Image Later: Two-Stage Scene Text Image Super-Resolution with Glyph Structure Guidance}

\author{%
Minxing Luo$^{1,\space2,\space*}$, 
Linlong Fan$^{2,\space*}$, 
Qiushi Wang$^{2,\space3}$, 
Ge Wu$^{1}$, 
Yiyan Luo$^2$,
Yuhang Yu$^2$,\\ 
Jinwei Chen$^2$, 
Yaxing Wang$^{1,\space4}$, 
Qingnan Fan$^{2,\space\dagger}$, 
Jian Yang$^{1,\space5,\space\dagger}$ \\
\small{$^1$PCA Lab, VCIP, College of Computer Science, Nankai University~~} 
\small{$^2$vivo BlueImage Lab, vivo Mobile Communication Co., Ltd.~~}
\\ 
 \small{$^3$SDS, CUHK, Shenzhen~~} \small{$^4$NKIARI, Shenzhen Futian~~}
 \small{$^5$PCA Lab, School of Intelligence Science and Technology, Nanjing University~~}
}

\begin{document}
\twocolumn[{%
\renewcommand\twocolumn[1][]{#1}%
\maketitle
\vspace{-30pt}
\includegraphics[width=0.95\linewidth]{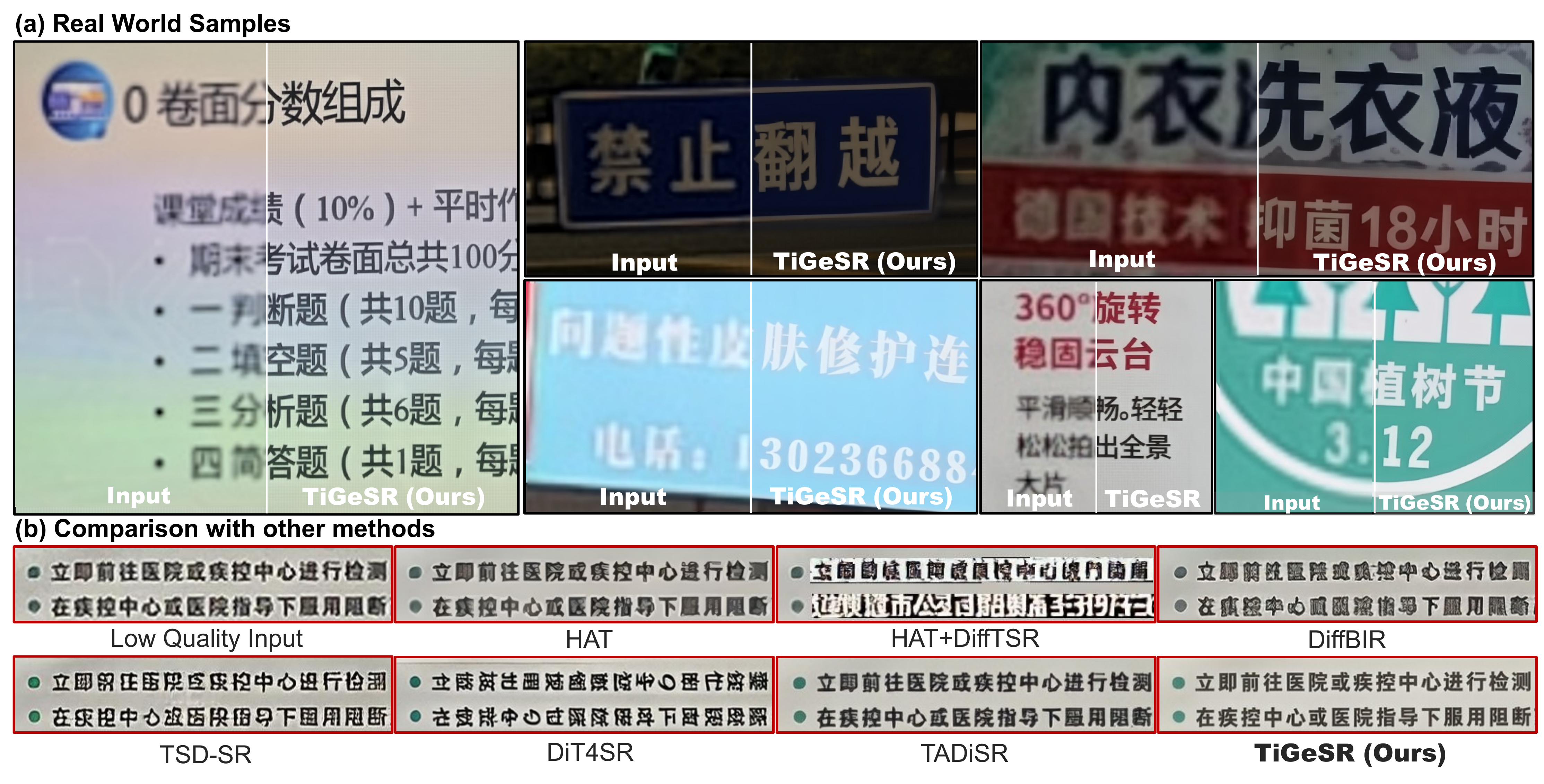}
\vspace{-15pt}
\centering
\captionof{figure}{We present \textbf{TiGeSR} (\textbf{T}ext–\textbf{i}mage \textbf{G}uid\textbf{e}d \textbf{S}uper-\textbf{R}esolution), a novel framework for scene text super-resolution. Its `text-first, image-later' paradigm ensures accurate glyph restoration and consistently high overall image fidelity and visual quality.} 
\vspace{0.6em}
\label{fig:teaser}
}]

\footnotetext[1]{Equal contribution.}
\footnotetext[2]{Corresponding author.}
\footnotetext{Work done during internship at vivo. Project page: \href{https://tony-lowe.github.io/TIGER_project_page/}{link}.}
\begin{abstract}

Current image super-resolution methods show strong performance on natural images but distort text, creating a fundamental trade-off between image quality and textual readability. 
To address this, we introduce \textbf{TiGeSR} (\textbf{T}ext–\textbf{i}mage \textbf{G}uid\textbf{e}d \textbf{S}uper-\textbf{R}esolution), a two-stage framework that breaks this trade-off through a \textit{``text-first, image-later"} paradigm. 
TiGeSR explicitly decouples glyph restoration from image enhancement: it first reconstructs precise text structures and uses them to guide full-image super-resolution. 
This ensures high fidelity and readability. 
For comprehensive training and evaluation, we present the UZ-ST (UltraZoom-Scene Text) dataset, the first Chinese scene text dataset with highest zoom of $\times 14.29$. 
Extensive experiments show TiGeSR achieves state-of-the-art performance, enhancing readability and image quality. 

\end{abstract}
    
\section{Introduction}
\label{sec:intro}

Scene Text Image Super-Resolution is a critical problem in computer vision for its vital role in downstream applications~\cite{DBLP:conf/iclr/LiC25,kil2023prestu,deshmukh2024textual,souibgui2023text}. 
It seeks to restore a high-quality super-resolution (SR) image from a degraded low-resolution (LR) input while preserving the text meaning.
Unlike general image super-resolution for natural scenes, where plausible texture synthesis or detail hallucination is acceptable and does not alter the semantic meaning of the image, visual texts exhibit extremely low tolerance to structural errors. 
This is especially evident in Chinese, where a minor stroke distortion or omission may entirely alter the meaning~\cite{yu2023chinese,li2023learning}.

In recent years, with the rapid development of generative models such as diffusion~\cite{ho2020denoising,dhariwal2021diffusion,saharia2022photorealistic,ramesh2022hierarchical,rombach2022high,chang2023muse,zhang2024artbank}, image super-resolution has increasingly relied on their strong generative priors to recover missing details from low-resolution (LR) inputs~\cite{wu2024seesr,yu2024scaling,lin2024diffbir,wu2025one,DBLP:conf/cvpr/DongFGWZCLZ25,duan2025dit4sr,hu2025text}. 
Although these approaches can effectively restore fine-grained natural textures (\eg, grass, leaves), they often distort text regions, turning them into gibberish as shown in Fig.\ref{fig:teaser} (b).
Because their primary goal is overall image enhancement, they tend to overlook text structure, resulting in a limited ability to preserve fine-grained glyph details.
This limitation is amplified for Chinese characters, due to their complex and diverse glyph designs and relatively low saliency in images. 
Consequently, models tend to under-represent and collapse glyphs into oversimplified, averaged forms, producing overlapping or distorted characters. 
Some researchers attempt to address this by processing only text regions~\cite{zhang2024diffusion,li2023learning}. 
While these methods improve readability notably, the absence of global background constraints often introduces style inconsistencies and block artifacts between text and background.
Ultimately, current approaches face a persistent trade-off between maintaining readability and ensuring high image quality.
\textit{Our key observation is that these goals need not be mutually exclusive if text and non-text are explicitly handled differently. 
Text structures can be reconstructed with dedicated mechanisms and then used to guide full-image restoration, ensuring coherent style without artifacts.}

Building on this insight, we introduce \textbf{TiGeSR} (\textbf{T}ext–\textbf{i}mage \textbf{G}uid\textbf{e}d \textbf{S}uper-\textbf{R}esolution), a progressive two-stage paradigm for scene text super-resolution built on the principle of \textit{``restoring text structure first, enhancing the whole image later." }
Unlike methods that rely on a single generative prior, TiGeSR explicitly separates the treatment of text and non-text regions: a diffusion-based local text refiner focuses on reconstructing fine-grained stroke geometry in text regions, ensuring glyph fidelity and structural consistency. 
The recovered text structures are then injected as conditional guidance into the subsequent full-image restoration stage, steering global super-resolution to harmonize text and background while suppressing artifacts and preserving overall visual quality.
\begin{table}[t]
\footnotesize
\caption{Dataset comparison. 
TextZoom~\citep{wang2020scene} lacks Chinese, CTR~\citep{yu2021benchmarking} lacks multi-line text, and Real-CE~\citep{ma2023benchmark} has mild degradation. UZ-ST covers all, enabling comprehensive evaluation. 
}
\vspace{-15pt}
\setlength{\tabcolsep}{1pt}
\label{tab:dataset}
\begin{center}
\begin{tabular}{lcccc} 
\toprule
Content
& TextZoom
& CTR
& Real-CE
& \bf UZ-ST (Ours)

\\
\midrule
Chinese characters
& \ding{55}
& \ding{51}
& \ding{51}
& \ding{51}
\\
Multi-line text
& \ding{55}
& \ding{55}
& \ding{51}
& \ding{51}
\\
Real-World Degradation($>\!\times 4$)
& \ding{55}
& \ding{55}
& \ding{55}
& \ding{51}
\\

\bottomrule

\end{tabular}
\end{center}
\vspace{-20pt}
\end{table}

Moreover, existing scene text super-resolution datasets~\cite{wang2020scene,yu2021benchmarking,ma2023benchmark} are limited to mild degradation ($\leq\!\times4$), which fails to reflect real-world long-distance scenarios and restricts models from learning robust low- to high-resolution mappings.
To address the issue of data scarcity and enable comprehensive evaluation~\cite{wang2020scene,ma2023benchmark}, we introduce the \textbf{UZ-ST} (UltraZoom-Sence Text) benchmark dataset, the first scene text dataset that contains a maximum zooming mode of $\times 14.29$, providing extra challenging scenarios for the field.
Content differences are outlined in \cref{tab:dataset}.
It includes high-quality 5,036 LR–HR pairs captured at multiple focal lengths (ranging from 14mm to 200mm), with 49,675 text lines in total.
Each pair comes with detailed annotations, including detection boxes and text transcripts, to support both training and evaluation.
The dataset includes diverse scenarios such as shop signs, posters, and documents. 
It also covers varying lighting conditions, including daylight, indoor lighting, and nighttime, offering a challenging yet practical benchmark for the field.
To ensure reliable alignment across focal lengths, we adopt a coarse-to-fine cascade alignment strategy combined with manual filtering: coarse alignment is performed via global geometric transformation, and residual misalignments are corrected through local refinement.
This process achieves accurate LR–HR alignment, thereby ensuring both the quality of the dataset and the reliability of our experiments. The main contributions are summarized as follows:
\begin{enumerate}
    \item We propose TiGeSR, the first two-stage scene text super-resolution framework with a \textit{`text-first, image-later'} paradigm that decouples glyph restoration from image enhancement, improving readability and visual quality.
    \item We introduce UZ-ST, the first scene text benchmark with maximum zoom of $\times$14.29, offering aligned, richly annotated LR–HR pairs for thorough evaluation under challenging real-world conditions.
    \item Extensive experiments on Real-CE and UZ-ST show our method outperforms prior state-of-the-art models, especially in preserving text structure.
\end{enumerate}

\section{Related Works}

\begin{figure*}[t]
\begin{center}
\includegraphics[width=0.95\linewidth]{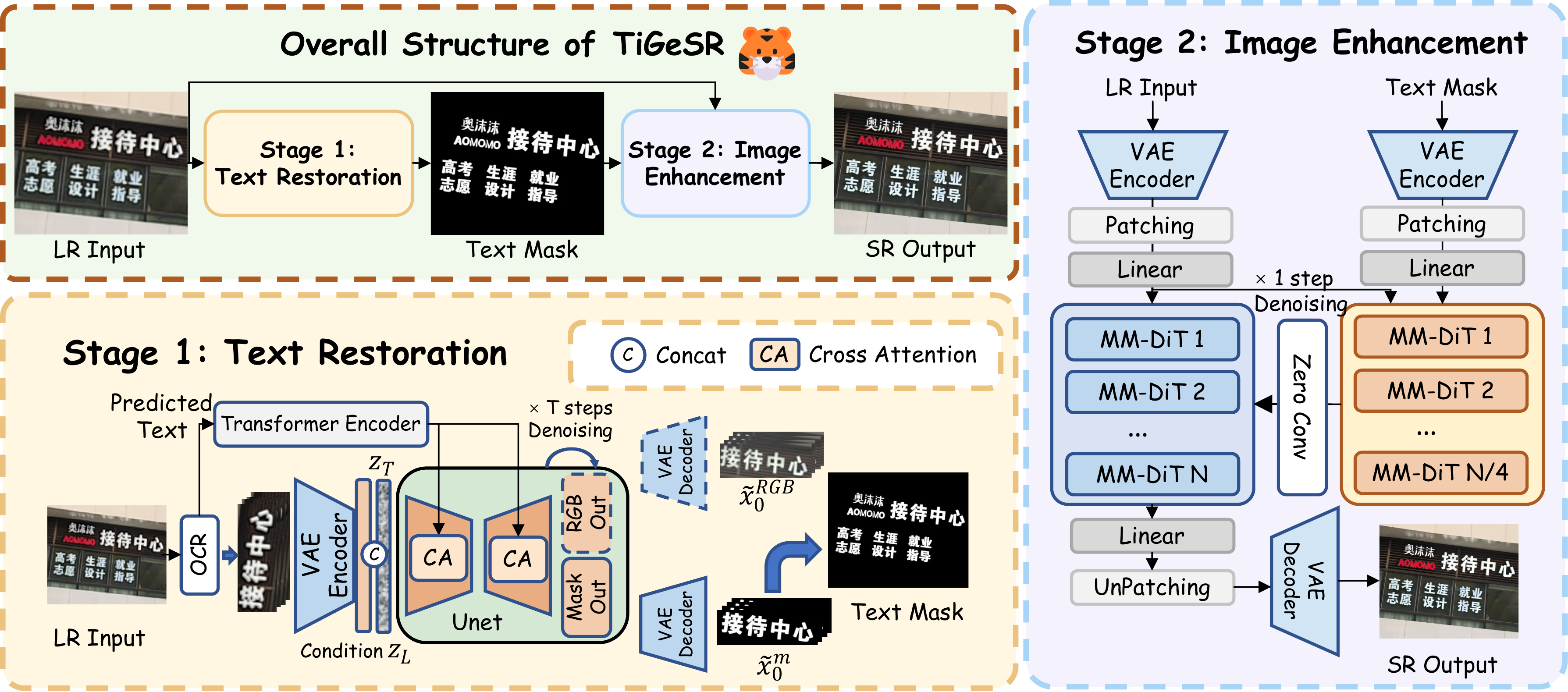}
\end{center}
\vspace{-15pt}
\caption{The framework of TiGeSR, which includes the Text Restoration stage (stage 1) and the Image Enhancement stage (stage 2). 
Stage 1 recovers accurate glyph structures from text regions. 
Stage 2 uses them to guide full-image restoration for coherent text and background.
}
\vspace{-15pt}
\label{fig:pipeline}
\end{figure*}

\noindent \textbf{Real-World Image Super-Resolution.}~
Real-world image super-resolution (Real-SR) aims to reconstruct high-resolution (HR) images from low-resolution (LR) inputs degraded under uncontrolled real-world conditions.
Early GAN-based methods, such as BSRGAN~\cite{zhang2021designing} and Real-ESRGAN~\cite{wang2021real}, synthesize degradations through random combinations of known distortions and use adversarial training for restoration.
While they generate natural-looking images, their instability and insensitivity to fine details limit their ability to recover structural elements, especially text.
Recent work introduces diffusion models into Real-SR~\cite{yue2023resshift,wang2024sinsr}, improving perceptual quality but still struggling with complex structures like scene text.
StableSR~\cite{wang2024exploiting} and DiffBIR~\cite{lin2024diffbir} apply ControlNet~\cite{zhang2023adding} to condition generation on LR inputs, while PASD~\cite{yang2024pixel} and SeeSR~\cite{wu2024seesr} incorporate high-level semantics to enhance fidelity. 
SUPIR~\cite{yu2024scaling} scales training with large image-text pairs and introduces degradation-robust encoders.
Other approaches, including OSEDiff~\cite{wu2025one} and TSD-SR~\cite{dong2025tsd}, directly apply the diffusion process on LR images and distill models for one-step sampling. 
DiT-SR~\cite{cheng2025effective}, DreamClear~\cite{ai2024dreamclear}, and DiT4SR~\cite{duan2025dit4sr} adopt diffusion transformers (DiT) for Real-SR. 
Despite enhancing perceptual fidelity, these methods often overlook accurate text structures.
TADiSR~\cite{hu2025text} addresses Chinese scene text super-resolution by using Kolors as the base model and aggregating cross-attention maps for text structure supervision. 
However, constrained by the resolution of cross-attention, it struggles to restore small or severely degraded text.

\noindent \textbf{Text Image Super-Resolution.}~Text image super-resolution (Text-SR) restores textual content from cropped images of isolated words or text lines.
Early methods~\cite{dong2015boosting} apply general SR architectures such as SRCNN~\cite{dong2014learning} to enhance OCR performance.
TextSR~\cite{wang2019textsr} introduces GANs with a recognition loss, while PlugNet~\cite{mou2020plugnet} and STT~\cite{chen2021scene} jointly train SR and recognition modules for more discriminative features.
TSRN~\cite{wang2020scene} releases the TextZoom dataset and employs an edge-aware module to preserve text details, while TATT~\cite{ma2022text} uses global attention for irregular layouts.
Methods like MTDM~\cite{wang2024multi} inpaint text but fail on complex scripts due to primitive architectures.
MARCONet~\cite{li2023learning} employs StyleGAN priors and a glyph structure codebook for realistic text reconstruction. 
DiffTSR~\cite{zhang2024diffusion} employs latent diffusion to separately denoise text and text-image components.
Despite progress in text structure restoration, existing methods lack global background constraints, causing style inconsistencies and block artifacts between text and background.

\section{Methodology}

\subsection{Architecture Overview}
\label{sec:ao}

Current image super-resolution methods focus on enhancing overall image quality but often fail to accurately preserve glyph structures, leading to distorted text in super-resolved images, as shown in \cref{fig:teaser} (b).
Conversely, text image super-resolution methods improve text readability but do not retain the global semantic information of the background, causing incoherence between text and non-text regions.
To leverage the strengths of both approaches, we introduce the TiGeSR framework.
As illustrated in \cref{fig:pipeline}, the TiGeSR framework is composed of 2 stages, the \textbf{Text Restoration stage} and the \textbf{Image Enhancement stage}. 
In the text restoration stage, we extract the text regions from the LR input $x_{L} \in \mathbb{R}^{H \times W \times C}$ and feed them into the glyph structure restoration model to restore the text structure based on the text region of the LR input and the predicted text. 
We then reassemble the text structures to their original positions to obtain a text mask $\hat{x}_{m} \in \mathbb{R}^{H \times W \times C}$. 
In the image enhancement stage, the text mask and LR input are then processed by a ControlNet-like network to obtain the enhanced SR output $\hat{x}_{H} \in \mathbb{R}^{H \times W \times C}$.
The following sections detail each stage.

\subsection{Text Restoration stage}
\label{sec:TR}


\begin{figure*}[t]
\begin{center}
\includegraphics[width=1.0\linewidth]{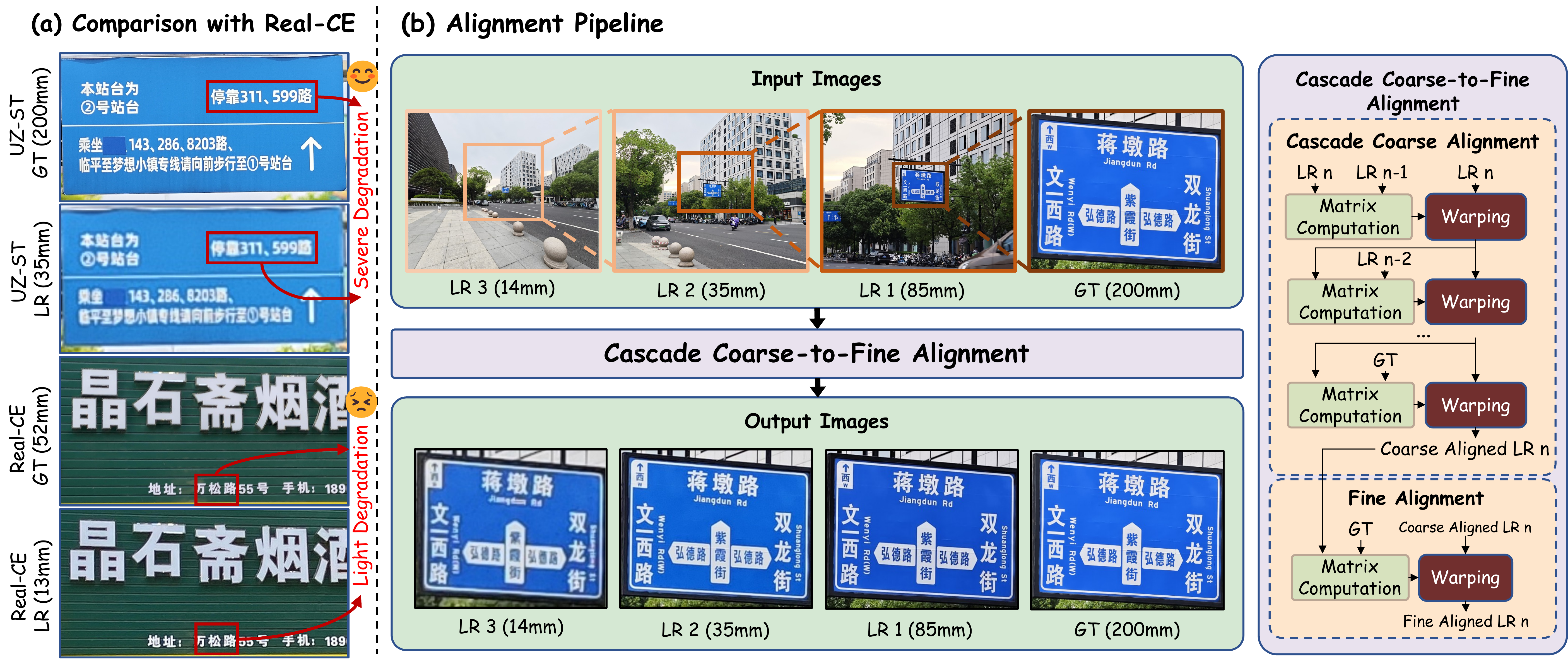}
\end{center}
\vspace{-18pt}
\caption{
Overview of UZ-ST (UltraZoom-Scene Text). 
(a) Real-CE LRs show only mild degradation (\textcolor{red}{red box}), while UZ-ST LRs exhibit stronger degradation (\textcolor{red}{red box}), enabling a more comprehensive evaluation.
(b) Coarse-to-fine alignment: images are sorted by focal length, each warped to the next higher-focal neighbor using an estimated homography matrix, then refined to the 200 mm GT.
}
\vspace{-15pt}
\label{fig:difference}
\end{figure*}

\noindent \textbf{Text Restoration Pipeline.}
Existing methods like Hi-SAM~\citep{ye2024hi} work well on clean high-resolution text but fail on the incomplete, distorted text in low-resolution images. 
To address this, we propose a region-level text restoration pipeline shown in \cref{fig:pipeline} built on~\cite{rombach2022high} to recover glyph structures $x_{m} \in \mathbb{R}^{H \times W \times C}$. 
An OCR detector first localizes text regions $\{x^{0}_{L},...,x^{N-1}_{L}\}$ and extracts their contents $\{y^{0},...,y^{N-1}\}$ as semantic conditions.
Each region $\Tilde{x}_{L}$ is encoded by VAE~\citep{kingma2013auto} into $\Tilde{z}_{L} \in \mathbb{R}^{h \times w \times c}$, concatenated with noise $z_{T} \in \mathbb{R}^{h \times w \times 2c}$, and iteratively denoised by a UNet $\epsilon_{\theta}$ into two branches: $z^{RGB}_{t-1} \in \mathbb{R}^{h \times w \times c}$ for appearance and $z^{m}_{t-1} \in \mathbb{R}^{h \times w \times c}$ for structure, which are merged as $z_{t-1}$. 
The text content $y$ is embedded into $c_{te}$ and fused into $z_{t}$ via cross-attention to guide structure recovery. After $T$ denoising steps, the mask branch output $z^{m}_{0}$ is decoded into $\Tilde{x}_{m}$, and all restored regions are assembled into the final text mask $\hat{x}_{m}$. 
Focusing on real text reconstruction enables the model to capture glyph structures without non-text interference and reduces sensitivity to text saliency.

\noindent \textbf{Training Strategy.}~The scarcity of real-world segmentation masks for degraded text forces reliance on synthetic data, yet its artificial degradation limits generalization. 
Annotating real degraded samples, however, is labor-intensive and costly.
To marry synthetic precision with real-world degradation, we propose a two-phase training strategy. 
In Phase 1, both synthetic and real data are included in training. 
This allows the model to generate text masks from the LR image and capture real-world degradation patterns. 
However, noisy masks in real data degrade mask quality.
To address this, we freeze the RGB out block and the mask out block and train the UNet with only synthetic data to refine the quality of output text masks in Phase 2. 
The training objective is described as follows:
\begin{equation}
    \mathcal{L}=\lambda_{td}\mathcal{L}_{td}+\lambda_{Seg}\mathcal{L}_{Seg},
\end{equation}
where $\mathcal{L}_{td}$, $\mathcal{L}_{Seg}$, $\lambda_{td}$, and $\lambda_{Seg}$ denote the text-control diffusion loss, segmentation-oriented loss, and their corresponding hyperparameters. The text-control diffusion loss is formulated as follows:
\begin{equation}
\mathcal{L}_{\epsilon}=\mathbb{E}_{\textbf{\textit{z}}_0,\Tilde{\textbf{\textit{z}}}_L, \textbf{\textit{c}}_{te}, \textbf{\textit{t}}, \epsilon\sim\mathcal{N}(0,1)}\left[\|\epsilon-\epsilon_{\theta}(\textbf{\textit{z}}_t, \Tilde{\textbf{\textit{z}}}_L, \textbf{\textit{c}}_{te}, \textbf{\textit{t}})\|_{2}^{2}\right].
\end{equation}
To improve the quality of the text mask, we propose a segmentation-oriented loss. Let $\mathcal{\varepsilon}_t$ denote the noise predicted by the denoiser network $\epsilon_{\theta}$.
Following ~\cite{ho2020denoising}, $z^{m}_0$ can be estimated by combining the time step $t$ with the noisy latent image $z^{m}_t$. 
This estimate is subsequently passed through the VAE decoder to obtain an approximate reconstruction of the original input text mask, denoted as $x'^{m}_{0}$.
In this way, text mask generation can be supervised at the pixel level. 
We combine Mean Squared Error (MSE), Focal, and Dice Losses, commonly used in segmentation tasks~\citep{ye2024hi}, to compare $x'^{m}_{0}$ with the original image $x^{m}_{0}$:
\begin{equation}
\begin{aligned}
     \mathcal{L}_{Seg} = \| x'^{m}_{0} - x^{m}_{0} \|^2_2 + \lambda_{Focal} \text{FocalLoss}(x'^{m}_{0},x^{m}_{0}) \\
     + \lambda_{Dice} \text{DiceLoss}(x'^{m}_{0}, x^{m}_{0}),
\end{aligned}
\end{equation}
where $\lambda_{Focal}$ and $\lambda_{Dice}$ are their respective balancing coefficients.
This combination forms a novel text-first paradigm that equips the model with both structural accuracy and real-world generalization, laying the foundation for the subsequent full-image super-resolution stage.


\subsection{Image Enhancement stage}

\noindent \textbf{Image Enhancement pipeline.}~To effectively take advantage of the generated glyph structure $\hat{x}_m$ and enhance the quality of LR $x_L$, we adopt a ControlNet~\citep{zhang2023adding} network $\epsilon_{\phi}$. 
As illustrated in~\cref{fig:pipeline}, after getting their latent representation $\hat{z}_m$ and $z_{L}$, we use the network to denoise the $z_{L}$ only at the specific timestep $t$, using the null-text embedding $c_{Null}$. 
The output super-resolved latent representation $\hat{z}_H$ will be denoted as:
\begin{equation}
    \hat{z}_H = z_L - \sigma_t \epsilon_{\phi}(z_{L},\hat{z}_{m},t,c_{Null}),
\end{equation}
where $\sigma_{t}$ is a scalar determined by the predefined diffusion time step. 
With the text restoration stage, this image enhancement pipeline forms a progressive `text-first, image after' paradigm. Our design injects structure-aware control into the generative process, allowing the network to enhance global image quality while preserving text structures.

\noindent \textbf{Training Strategy.}~For the quality of image super-resolution, we constrain the reconstruction loss between the predicted high-resolution image $\hat{x}_{H}$ decoded from $\hat{z}_{H}$ by $\varepsilon$ and the ground truth high-resolution image $x_{H}$ using a weighted sum of MSE and LPIPS losses:
\begin{equation}
    \mathcal{L}_{img} = \lambda_{l2} \| x_H - \hat{x}_H \|^{2}_2 + \lambda_{LPIPS} \text{LPIPS}(x_H,\hat{x}_H),
\end{equation}
where $\lambda_{l2}$ and $\lambda_{LPIPS}$ are balancing coefficients for different loss terms. 
To enhance glyph control and emphasize the glyph structure, we extract the text boundaries using Sobel operators~\citep{roberts1987digital}. This edge loss is expressed as:
\begin{equation}
    \mathcal{L}_{edge} = \|  \text{Sobel} (x_H) - \text{Sobel} (\hat{x}_H) \|^{2}_2.
\end{equation}
We combine the two loss terms and use $\lambda_{edge}$ as the balancing coefficient in the complete loss function for stage 2:
\begin{equation}
    \mathcal{L} = \mathcal{L}_{img} + \lambda_{edge} \mathcal{L}_{edge}.
\end{equation}

The combination of glyph-aware ControlNet conditioning and edge-constrained training objectives enables our model to preserve stroke integrity while harmonizing text and background appearance.




\section{Dataset and Benchmark}
\label{sec:4_dataset}

Existing datasets like \citet{ma2023benchmark} offer only subtle degradation, as shown in \cref{fig:difference} (a), making them insufficient for evaluating model robustness.
To address the lack of both challenging and publicly available datasets tailored for scene text image super-resolution, especially for Chinese text, we introduce \textbf{UZ-ST}, a challenging real-world benchmark.
It is collected using ViVO X200 Ultra equipped with four fixed focal lengths (14 mm, 35 mm, 85 mm, 200 mm), enabling image pairs with a maximum zoom of $\times 14.29$.
It includes diverse scenes—street views, book covers, advertisements, menus, and posters—captured under varied lighting, with all text lines manually annotated.
However, such zoom introduces severe misalignment that breaks pixel-wise optimization methods like~\cite{cai2019toward}, originally designed for moderate $\times 2$ to $\times 4$ zoom.
To address this, we design a \textbf{Cascade Coarse-to-fine Alignment} pipeline (see \cref{fig:difference} (b)): images are first sorted by focal length, and each low-quality image is sequentially aligned to its next higher-focal neighbor using optimization-based registration or feature-based extractors \cite{lowe1999object,ripe2025}, then refined against the 200 mm ground truth.
Finally, we manually filter or realign problematic cases to ensure precise alignment.

\begin{table*}[t]
\footnotesize
\caption{Evaluation results on image quality and text accuracy. 
Numbers in \textbf{bold} indicate the best performance.
TiGeSR performs best.}
\label{tab:MainTable}
\begin{center}
\setlength{\tabcolsep}{4pt}
\vspace{-15pt}
\begin{tabular}{lcccccc|cccccc}  
\toprule
\multirow{2}{*}{\bf Methods} & \multicolumn{6}{c}{\bf Real-CE} & \multicolumn{6}{c}{\bf UZ-ST} \\
~ 
&  \multicolumn{1}{c}{\bf PSNR$\uparrow$}  
&  \multicolumn{1}{c}{\bf SSIM$\uparrow$ }
&  \multicolumn{1}{c}{\bf LPIPS$\downarrow$ }
&  \multicolumn{1}{c}{\bf DISTS$\downarrow$ }
&  \multicolumn{1}{c}{\bf FID$\downarrow$ }
&  \multicolumn{1}{c}{\bf OCR-A $\uparrow$}
&  \multicolumn{1}{c}{\bf PSNR$\uparrow$}  
&  \multicolumn{1}{c}{\bf SSIM$\uparrow$ }
&  \multicolumn{1}{c}{\bf LPIPS$\downarrow$ }
&  \multicolumn{1}{c}{\bf DISTS$\downarrow$ }
&  \multicolumn{1}{c}{\bf FID$\downarrow$ }
&  \multicolumn{1}{c}{\bf OCR-A $\uparrow$ }
\\ 
\midrule
Real-ESRGAN &
  22.30 &
  0.787 &
  0.239 &
  0.188 &
  53.60 
  & 56.0\%
 & 23.99 &
  0.790 &
  0.248 &
  0.194 &
  30.60
 & 37.1\%
  \\
HAT &
  23.61 &
  0.830 &
  0.214 &
  0.176 &
  51.16 
   & 56.6\%
  & 25.17 &
  0.815 &
  0.249 &
  0.198 &
  30.12 
  & 37.9\%
  \\
MARCONet &
  21.89 &
  0.785 &
  0.238 &
  0.150 &
  52.97 
    & 55.0\%
  & 22.13 &
  0.768 &
  0.306 &
  0.205 &
  34.28 
  & 33.4\%
  \\
SeeSR &
  23.59 &
  0.822 &
  0.195 &
   0.169 &
  43.75
  & 37.4\%
  & 23.64 &
  0.788 &
  0.219 &
  0.184 &
  26.73 
  & 22.8\% 
  \\
SupIR &
  21.78 &
  0.723 &
  0.310 &
  0.198 &
  44.94
  & 27.8\%
  & 23.62 &
  0.754 &
  0.308 &
  0.207 &
  27.91 
  & 23.6\%
  \\
DiffTSR &
  22.10 &
  0.768 &
  0.278 &
  0.168 &
  44.91
  & 44.1\%
    & 22.41 &
  0.767 &
  0.300 &
  0.189 &
  31.25 
  & 31.7\%
  \\
DiffBIR &
   22.44 &
   0.747 &
    0.260 &
    0.201 &
   46.44
   & 37.6\%
   & 23.67 &
   0.724 &
   0.262 &
   0.197 &
   23.10  
   & 26.6\%
  \\
OSEDiff &
  21.86 &
  0.771 &
  0.197 &
  0.127 &
  41.00  
  & 26.8\%
  & 25.07
 & 0.819
 & 0.201
 & 0.169
 & 20.53
 & 28.9\%

  \\
DreamClear &
  22.47 &
  0.772 &
  0.216 &
  0.157 &
  38.97
  & 50.2\%
  & 24.10 
& 0.773 
& 0.238 
& 0.191 
& 21.75 
& 21.4\%
  \\
TSD-SR &
  21.43 &
  0.754&
  0.220&
  0.175&
  47.21  
  & 34.9\%
  & 22.79 
  & 0.757&
  0.207&
  0.194&
  24.08  
  & 26.6\% 
  \\
DiT4SR &
   20.54 &
   0.764 &
   0.268 &
   0.186 &
   49.79 
   & 29.2\%
   & 23.16 
   & 0.767 
   & 0.215 
   & 0.159 
   & 20.58 
   & 23.7\%
  \\
TADiSR &
   23.83 &
   0.790 &
   0.286 &
   0.154 &
   44.42 
   & 64.7\%
   & 24.61 
   & 0.796 
   & 0.203 
   & 0.160 
   & 36.61
   & 36.6\% 
  \\
\rowcolor{gray!30} \bf TiGeSR (Ours) &
   \textbf{24.12}  
   & \textbf{0.839} &
   \textbf{0.164} &
   \textbf{0.125} &
   \textbf{38.72}  
   & \textbf{67.3\%}
   & \textbf{25.48} 
   & \textbf{0.830}
   & \bf{0.196}
   & \bf{0.156}
   & \textbf{20.01}   
   & \textbf{43.0\%}
  \\ 
\bottomrule

\end{tabular}
\end{center}
\vspace{-15pt}
\end{table*}
\begin{table*}[t]
\footnotesize
\caption{Evaluation of image quality on text regions and text accuracy compared to LR ($\Delta$ OCR-A).
TiGeSR achieves the best performance.}
\label{tab:region}
\begin{center}
\setlength{\tabcolsep}{1.5pt}
\vspace{-15pt}
\begin{tabular}{lccccc|ccccc} 
\toprule
\multirow{2}{*}{\bf Methods} & \multicolumn{5}{c}{\bf Real-CE} & \multicolumn{5}{c}{\bf UZ-ST} \\
~ 
&  \multicolumn{1}{c}{\bf PSNR$_{cr}$$\uparrow$}
&  \multicolumn{1}{c}{\bf SSIM$_{cr}$ $\uparrow$ }
&  \multicolumn{1}{c}{\bf LPIPS$_{cr}$ $\downarrow$}
&  \multicolumn{1}{c}{\bf DISTS$_{cr}$ $\downarrow$}
&  \multicolumn{1}{c}{\bf $\Delta$ OCR-A $\uparrow$}
&  \multicolumn{1}{c}{\bf PSNR$_{cr}$ $\uparrow$}
&  \multicolumn{1}{c}{\bf SSIM$_{cr}$ $\uparrow$}
&  \multicolumn{1}{c}{\bf LPIPS$_{cr}$ $\downarrow$}
&  \multicolumn{1}{c}{\bf DISTS$_{cr}$ $\downarrow$}
&  \multicolumn{1}{c}{\bf $\Delta$ OCR-A $\uparrow$}
\\ 
\midrule
Real-ESRGAN 
& 21.71
& 0.824
& 0.257 
& 0.224
& -8.8\%
& 21.25
  & 0.786
  & 0.302
  & 0.266
  & -4.7\%
  \\
HAT 
& 23.03
& 0.862
& 0.249
& 0.239
& -8.2\%
& 22.18
& 0.813
& 0.291
& 0.276
& -3.9\%
  \\

MARCONet 
& 20.68
& 0.786
& 0.267
& 0.223
& -9.8\%
& 17.48
& 0.690
& 0.529
& 0.366
& -8.3\%

  \\
SeeSR 
& 22.74
& 0.844
& 0.237
& 0.212
& -27.4\%
& 20.45
& 0.767
& 0.297
& 0.241
& -18.9\%
  \\
 SupIR 
& 19.96
& 0.778
& 0.326
& 0.263
& -37.0\%
& 20.10
& 0.756
& 0.344
& 0.287
& -18.2\%
  \\

 DiffTSR 
& 20.46
& 0.818
& 0.275
& 0.239
& -20.7\%
& 17.00
& 0.665
& 0.452
& 0.313
& -10.0\%
  \\
 DiffBIR 
& 21.20
& 0.792
& 0.278
& 0.224
& -27.2\%
& 20.70
& 0.757
& 0.302
& 0.245
& -15.2\%
  \\
 OSEDiff 
& 19.38
& 0.768
& 0.269 
& 0.208
& -38.0\%
& 21.43
& 0.798
& 0.276
& 0.263
& -12.8\%
  \\
 DreamClear 
& 22.74
& 0.845
& 0.190
& 0.168
& -14.6\%
& 20.50
& 0.768
& 0.317
& 0.276
& -20.3\%
  \\

TSD-SR
& 19.33
& 0.763
& 0.279
& 0.234
& -29.9\% 
& 19.42
& 0.748
& 0.296
& 0.249
& -15.1\% 
  \\
 DiT4SR 
& 17.95
& 0.738
& 0.317
& 0.244
& -35.6\%
& 19.73
& 0.760
& 0.273
& 0.216
& -18.0\%
  \\

TADiSR 
& 23.39
& 0.855
& 0.253
& 0.258
& -1.0\% 
& 21.59
& 0.799
& 0.360
& 0.336
& -5.1\% 
  \\

\rowcolor{gray!30} \bf TiGeSR (Ours) 
 & \textbf{23.43}
 & \textbf{0.864}
 & \textbf{0.173}
 & \textbf{0.167}
 & \textbf{+2.5\%} 
 & \textbf{22.22}
 & \textbf{0.814}
 & \textbf{0.228}
 & \textbf{0.212}
 & \textbf{+1.3\%}
\\
\bottomrule
\end{tabular}
\end{center}
\vspace{-21pt}
\end{table*}

Through meticulous annotation and alignment, we obtained 5,036 image pairs with 49,675 text lines.
We treat images captured under 200 mm focal lengths as GT, yielding 1,439, 1,798, and 1,799 pairs for the $\times14.29$, $\times5.71$, and $\times2.35$ zoom modes, respectively.
Among them, we randomly select 470, 589, and 581 pairs for evaluation under each zoom mode.
Each image pair contains one or more text lines.
These evaluation sets allow us to assess model performance in more complex and challenging scenarios.

For the UZ-ST-benchmark, we utilize 5 evaluation metrics to examine the quality of image super-resolution on the scale of the full image.
Firstly, we adopt Peak Signal-to-Noise Ratio (PSNR), Structural Similarity Index Measure (SSIM)~\citep{wang2019textsr}, Learned Perceptual Image Patch Similarity (LPIPS)~\citep{zhang2018unreasonable}, Deep Image Structure and Texture Similarity (DISTS)~\citep{ding2020image}, and Fréchet Inception Distance (FID)~\citep{heusel2017gans} for evaluation of image quality. 
Specifically, PSNR and SSIM are computed in the pixel space to quantify low-level reconstruction fidelity, LPIPS and DISTS are computed in the feature space to assess perceptual similarity, and FID is employed to evaluate the distributional discrepancy between generated and real images.
However, these metrics do not directly reflect the quality or accuracy of super-resolved text. To address this, we crop annotated text regions from the images and compute PSNR, SSIM, LPIPS, and DISTS on these cropped regions, denoted as $\text{PSNR}{cr}$, $\text{SSIM}{cr}$, $\text{LPIPS}{cr}$, and $\text{DISTS}{cr}$.
For text accuracy, we apply an OCR model to recognize the text and compare its outputs with the ground-truth annotations using the Levenshtein ratio~\citep{yujian2007normalized}:
\begin{equation}
    \begin{aligned}
    \text{OCR-A} = \frac{\text{Len}(s_{pred})+\text{Len}(s_{gt})-\text{Dist}(s_{pred},s_{gt})}{\text{Len}(s_{pred})+\text{Len}(s_{gt})},
    \end{aligned}
\end{equation}
where $s_{pred}$ stands for the predicted text sequence, $s_{gt}$ denotes annotated ground-truth text sequence, and $\text{Dist}(\cdot, \cdot)$ is the Levenshtein distance between text sequences.

\begin{figure*}[t]
\begin{center}
\includegraphics[width=1.0\linewidth]{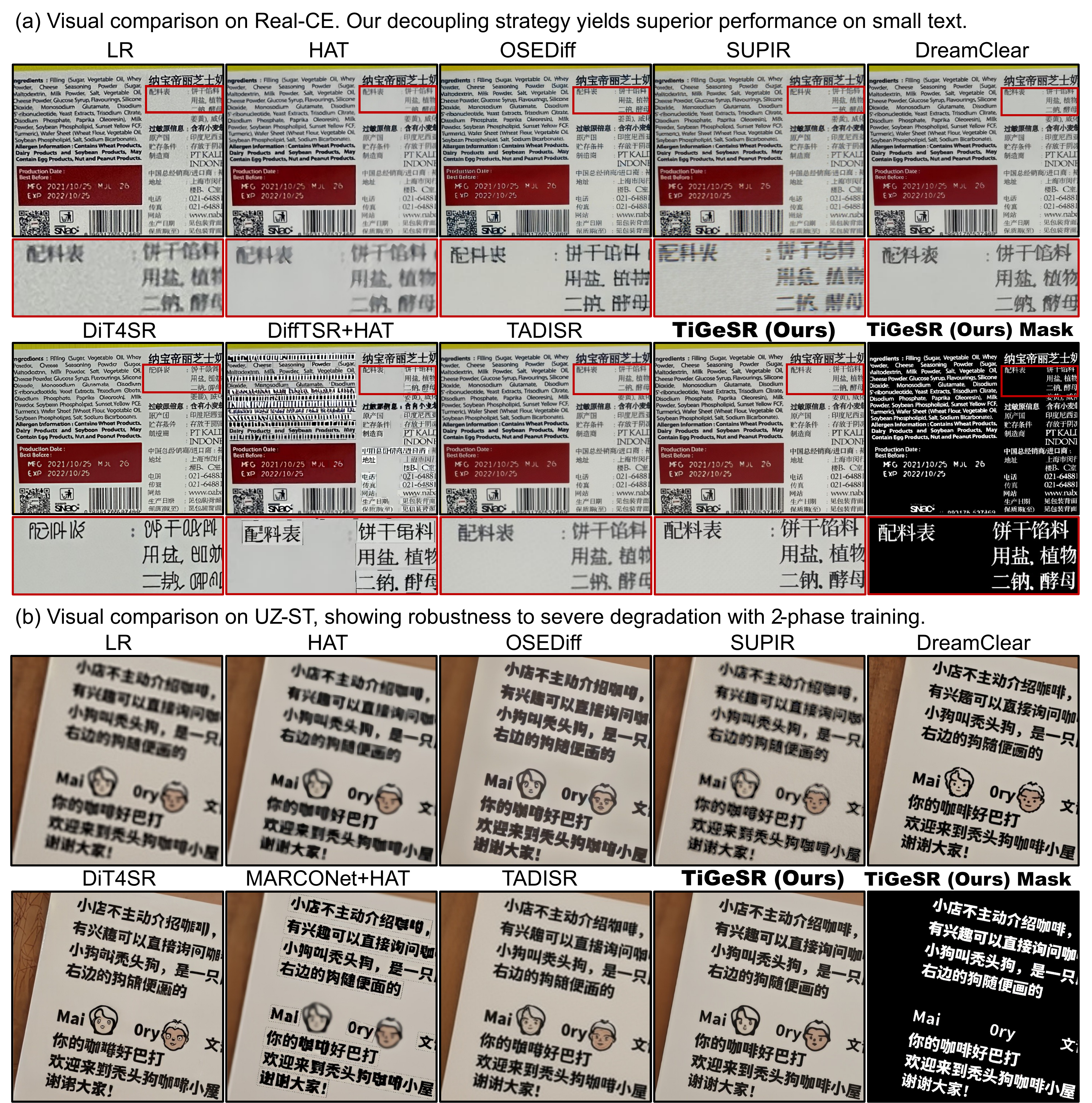}
\end{center}
\vspace{-15pt}
\caption{Qualitative Evaluation on Real-CE and UZ-ST.}
\vspace{-15pt}
\label{fig:ultrazoom}
\end{figure*}

\section{Experiment}
\subsection{Implementation Details}
\label{sec:Imple}

We combine synthetic data (built upon LSDIR~\cite{li2023lsdir} with text rendered via LBTS~\cite{tang2023scene} and Real-ESRGAN~\cite{wang2021real} degradation) and real data from Real-CE~\cite{ma2023benchmark} and UZ-ST. Misaligned pairs in Real-CE are filtered and reannotated, yielding 337 training and 188 testing pairs. Stage 1 uses cropped text regions and is trained using an IDM-based architecture~\cite{zhang2024diffusion} with combined segmentation and reconstruction losses. Stage 2 is based on Stable Diffusion 3.5~\cite{esser2024scaling} and employs a tile-based inference strategy~\cite{yu2024scaling,hu2025text}; it is pretrained on synthetic data and fine-tuned on real data. For more details, see the \cref{Appen:Implement}.

\subsection{Comparison Results}

\subsubsection{Quantitative Results}
We evaluate existing competing methods, including GAN-based image super-resolution approaches such as Real-ESRGAN~\cite{wang2021real} and HAT~\cite{chen2023activating}; diffusion-based approaches such as SeeSR~\cite{wu2024seesr}, SupIR~\cite{yu2024scaling}, DiffBIR~\cite{lin2024diffbir}, OSEDiff~\cite{wu2025one}, DreamClear~\cite{ai2024dreamclear}, TSD-SR~\cite{dong2025tsd}, and DiT4SR~\cite{duan2025dit4sr}; as well as text-focused reconstruction methods such as MARCONet~\cite{li2023learning}, DiffTSR~\cite{zhang2024diffusion}, and TADiSR~\cite{hu2025text}. 
Evaluations are performed on the Real-CE Benchmark~\cite{ma2023benchmark} and the benchmark described in \cref{sec:4_dataset}. 
To ensure fairness, we fine-tune the released pre-trained models on the training sets of both benchmarks using the official code when available.
Following~\cite{hu2025text}, we integrate the outputs of MARCONet and DiffTSR with HAT-generated results to simulate real-world application scenarios and enable comprehensive full-image evaluation. 
We evaluate Real-CE at the hardest difficulty level ($\times 4$), while UZ-ST is evaluated across all difficulty levels—with average results reported here. 
Detailed per-level evaluations appear in the \cref{sec:Appendix:additional}.

As shown in \cref{tab:MainTable} and \cref{tab:region}
, TiGeSR outperforms competing methods on both the Real-CE and UZ-ST benchmarks in terms of image quality and text accuracy. 
TADiSR, limited by the resolution of its cross-attention mechanism, performs poorly under severe degradation and with small text. 
DiffTSR and MARCONet fail to effectively handle text backgrounds and struggle with text regions that have a large width-to-height aspect ratio. 
In contrast, our method restores images with high quality and fine-grained text, owing to the decoupling strategy.
It achieves OCR-A scores above 0.67 and 0.43, and demonstrates overall superiority in both pixel-level and perceptual accuracy.



\begin{figure*}[ht]
\begin{center}
\includegraphics[width=1.0\linewidth]{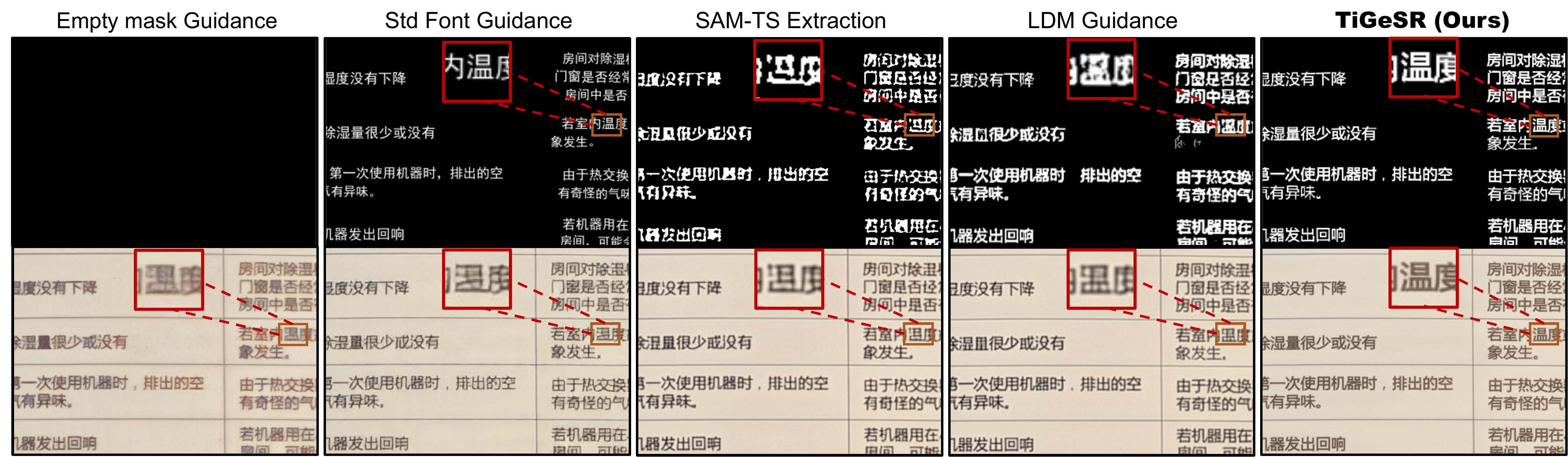}
\end{center}
\vspace{-15pt}
\caption{Qualitative Results of Ablation Study with stage 2 fixed as the baseline.}
\vspace{-15pt}
\label{fig:ablation}
\end{figure*}

\subsubsection{Qualitative Results}
\cref{fig:ultrazoom} shows visual comparisons on Real-CE and UZ-ST.
Real-world degradations cause noise, blurred strokes, and distortions in the input images. 
GAN-based methods reduce noise but fail to fix text structure, while diffusion-based SR methods, though powerful, lack text-structure guidance and often distort strokes, sometimes making text unreadable.

Methods tailored for text image super resolution (e.g., MARCONet and DiffTSR) exhibit limitations in processing long text sequences, often generating distorted or semantically meaningless results.
They also lack global semantic guidance, making it hard to blend text regions with backgrounds. 
As seen in \cref{fig:ultrazoom} (a), DiffTSR+HAT (red box) produces text colors inconsistent with the original LR and SR output of HAT. 
In \cref{fig:ultrazoom} (b), it even distorts the face near the word ``Ory”.
TADiSR works well on lightly degraded images but fails under severe degradation and performs poorly on small text due to cross-attention mask resolution limits.
Our method overcomes these issues, restoring high-quality glyph structures even in challenging cases.

\subsection{Ablation Study}

\begin{table}[t]
\footnotesize
\caption{Validation on the effectiveness of UZ-ST. Finetuning improves OCR-A, proving effectiveness. }
\label{tab:ablation1}
\setlength{\tabcolsep}{1pt}
\vspace{-15pt}

\begin{center}
\begin{tabular}{lccccccc} 
\toprule
\bf Methods & \bf PSNR$\uparrow$  & \bf SSIM$\uparrow$ & \bf LPIPS$\downarrow$ & \bf DISTS$\downarrow$ & \bf FID$\downarrow$ & \bf OCR-A$\uparrow$  \\
\midrule
OSEDiff w/o UZ-ST
 & 23.50
 & 0.791
 & \textbf{0.197 }
 & \textbf{0.162} 
 & 24.30
 & 22.8\%

\\
OSEDiff w/~~~UZ-ST
& \textbf{25.07}
 & \textbf{0.819}
 & 0.201
 & 0.169
 & \bf{20.53}
 & \textbf{28.9\%}
\\
\midrule
DiT4SR w/o UZ-ST  
& 22.58  
& 0.754 
& 0.252 
& 0.196 
& 26.93 
& 19.3\%  
\\
DiT4SR w/~~~UZ-ST 
& \textbf{23.17} 
& \textbf{0.767} 
& \textbf{0.215} 
& \textbf{0.159} 
& \textbf{20.58} 
& \textbf{23.7\%} 
\\
\midrule
Ours w/o UZ-ST     
& 22.96
& 0.782
& 0.341
& 0.253
& 33.78
& 40.0\%  \\
Ours w/~~~UZST 
& \textbf{25.48} 
& \textbf{0.830} 
& \textbf{0.196} 
& \textbf{0.156} 
& \textbf{20.01} 
& \textbf{43.0\%}  \\
\bottomrule

\end{tabular}
\end{center}
\vspace{-20pt}
\end{table}

\begin{table}[t]
\footnotesize
\caption{Ablation on OCR prediction.
The model remains strong with null or random OCR text, indicating limited reliance on OCR.}
\label{tab:ocr_ablation}
\begin{center}
\setlength{\tabcolsep}{0.5pt}
\vspace{-15pt}
\begin{tabular}{lcccccc} 
\toprule
\bf Methods
& \bf PSNR$\uparrow$
& \bf SSIM$\uparrow$
& \bf LPIPS$\downarrow$
& \bf DISTS$\downarrow$
& \bf FID$\downarrow$
& \bf OCR-A$\uparrow$
\\ 
\midrule
TADiSR
& 24.68
& 0.795
& 0.362
& 0.227
& 52.13
& 35.5\%
\\
Null Text
& 25.59
& \textbf{0.834}
& 0.188
& 0.151
& 28.93
& 40.4\%
\\
Random Text
& 25.60
& 0.833
& 0.188
& \textbf{0.150}
& 28.92
& 40.3\%
\\
\textbf{Predicted Text (Ours)}
& \textbf{25.63}
& \textbf{0.834}
& \textbf{0.185}
& \textbf{0.150}
& \textbf{28.82}
& \textbf{44.6\%}
\\
\bottomrule
\end{tabular}
\end{center}
\vspace{-25pt}
\end{table}

\noindent \textbf{Effectiveness of the training set.}
\cref{tab:ablation1} reports consistent OCR-A improvements across different architectures, including the UNet-based OSEDiff and the DiT-based DiT4SR.
This demonstrates that even models without explicit mechanisms for text structure modeling can still learn the mapping from low- to high-resolution text using our dataset, further validating its effectiveness.
Due to their lack of a dedicated mechanism for text structure modeling, they cannot outperform our methods.

\noindent \textbf{OCR Influence.}
To investigate how OCR prediction affects the accuracy of our framework, we set the predicted text by OCR in stage 1 as \textbf{null text} or \textbf{random text} and test them on the 35mm subset of UZ-ST.
As shown in \cref{tab:ocr_ablation}, even without the OCR model’s predicted text, our method attains a strong OCR-A score of 40.4\%, outperforming TAiDSR. 
This demonstrates that our model in stage 1 benefits from the inherent text structure in the LR input to restore text structure and is not overly dependent on the OCR model.

\noindent \textbf{Effectiveness of the TiGeSR component.}
We set stage 2 as the fixed baseline for image restoration and validate the effectiveness of our framework through 4 ablated variants: 
\begin{itemize}[leftmargin=10pt, itemindent=0pt,wide=\parindent]
    \item \textbf{Empty Mask Guidance.}
    As shown in \cref{tab:ablation1}, empty masks provide zero structure guidance, and the model can only rely on the LR for text restoration. Therefore, the OCR-A drops from 67.3\% to 59.5\%. 
    \item \textbf{Standard Font Guidance.}
    As shown in the second column of \cref{fig:ablation}, the standard font, while structurally correct, provides weak guidance due to style and position mismatches. 
    As a result, the restored text exhibits poor accuracy compared to TiGeSR, with the OCR-A dropping from 67.1\% to 55.3\%, as shown in \cref{tab:ablation}.
    \item \textbf{SAM-TS Extraction}~\cite{hu2025text,ye2024hi}. 
    While this improves visual quality by better aligning output masks with the LR image, SAM-TS can only extract distorted structures and can’t recover degraded text (\cref{fig:ablation}, third column), leading to suboptimal performance compared with TiGeSR, with OCR-A dropping from 0.671 to 0.579. 
    \item \textbf{LDM Guidance.}
    This variant uses a latent diffusion model conditioned on LR to restore text structure, achieving better accuracy by reconstructing or compensating for lost text. 
    However, without separate learning from synthetic and real-world data, it struggles with high-quality text structures in real-world scenarios. 
    As shown in \cref{fig:ablation}, fourth column, the masks remain noisy under real-world degradations, leading to lower accuracy than TiGeSR, with OCR-A dropping from 67.1\% to 60.1\% as reported in \cref{tab:ablation}.
\end{itemize}


Our method uses RGB output to jointly learn from real-world degradations and accurate synthetic masks, yielding high-quality masks and achieving the highest accuracy.

\begin{table}[t]
\footnotesize
\caption{Ablation study on Real-CE with stage 2 fixed as the baseline. From top to bottom, we compare the performance of using text masks rendered with a \textbf{standard font}, extracted using \textbf{SAM-TS}, reconstructed with \textbf{latent diffusion model} conditioned on LR, and reconstructed with our \textbf{text restoration pipeline}. Our pipeline faithfully restores glyph structures, yielding the highest accuracy.}
\vspace{-15pt}

\label{tab:ablation}
\setlength{\tabcolsep}{1pt}
\begin{center}
\begin{tabular}{lcccccc} 
\toprule
\bf Text Mask Settings & \bf PSNR$\uparrow$  & \bf SSIM$\uparrow$ & \bf LPIPS$\downarrow$ & \bf DISTS$\downarrow$ & \bf FID$\downarrow$ & \bf OCR-A$\uparrow$ \\
\midrule
Empty Mask
& 23.33
& 0.792
& 0.225
& 0.172
& 42.93
& 59.5\%
\\
Std Font Guidance
    & 23.31
    & 0.786
    & 0.249
    & 0.164
    & 41.83
    & 55.3\%
    \\
SAM-TS Extraction
    & 23.91
    & 0.835
    & 0.212
    & 0.147
    & 41.22
    & 57.9\%
    \\
LDM Guidance
    & 22.42 
    & 0.798 
    & 0.222
    & 0.147 
    & 43.08 
    & 60.1\%
    \\
\rowcolor{gray!30} \bf TiGeSR (Ours)&
   \textbf{24.12}  &
   \textbf{0.839} &
   \textbf{0.164} &
   \textbf{0.125} &
   \textbf{38.72}  &
   \textbf{67.3\%} 
  \\

\bottomrule

\end{tabular}
\end{center}
\vspace{-25pt}
\end{table}

\section{Conclusion}

In this paper, we delve into the extensively researched problem of scene text super-resolution. 
To address this challenge, we propose a novel approach, TiGeSR, that decouples glyph structure restoration from image enhancement. 
For the restoration of text structure, we propose the text restoration pipeline, which enables a 2-phase training strategy to fully take advantage of both synthetic and real-world data.
For the image enhancement, we propose the image enhancement pipeline, which effectively utilizes the glyph structure to restore both text fidelity
and image details coherently in full-image super-resolution.
In terms of training data and benchmark, we present the UltraZoom-ST, the first real-world scene text dataset with a maximum zooming of $\times 14.29$,
offering more challenging scenarios for the field. 
Extensive experiments on Real-CE and UZ-ST demonstrate the superiority of TiGeSR over existing methods.


\section*{Acknowledgement}
This work was supported by the NSFC under Grant Nos. 62361166670, U24A20330, Shenzhen Science and Technology Program (JCYJ20240813114237048), ``Science and Technology Yongjiang 2035" key technology breakthrough plan project (2024Z120), Chinese government-guided local science and technology development fund projects (scientific and technological achievement transfer and transformation projects) (254Z0102G).
{
    \small
    \bibliographystyle{ieeenat_fullname}
    \bibliography{main}

\begin{thebibliography}{59}
\providecommand{\natexlab}[1]{#1}
\providecommand{\url}[1]{\texttt{#1}}
\expandafter\ifx\csname urlstyle\endcsname\relax
  \providecommand{\doi}[1]{doi: #1}\else
  \providecommand{\doi}{doi: \begingroup \urlstyle{rm}\Url}\fi

\bibitem[Ai et~al.(2024)Ai, Zhou, Huang, Han, Chen, You, and Yang]{ai2024dreamclear}
Yuang Ai, Xiaoqiang Zhou, Huaibo Huang, Xiaotian Han, Zhengyu Chen, Quanzeng You, and Hongxia Yang.
\newblock Dreamclear: High-capacity real-world image restoration with privacy-safe dataset curation.
\newblock \emph{Advances in Neural Information Processing Systems}, 37:\penalty0 55443--55469, 2024.

\bibitem[Cai et~al.(2019)Cai, Zeng, Yong, Cao, and Zhang]{cai2019toward}
Jianrui Cai, Hui Zeng, Hongwei Yong, Zisheng Cao, and Lei Zhang.
\newblock Toward real-world single image super-resolution: A new benchmark and a new model.
\newblock In \emph{Proceedings of the IEEE/CVF international conference on computer vision}, pages 3086--3095, 2019.

\bibitem[Chang et~al.(2023)Chang, Zhang, Barber, Maschinot, Lezama, Jiang, Yang, Murphy, Freeman, Rubinstein, Li, and Krishnan]{chang2023muse}
Huiwen Chang, Han Zhang, Jarred Barber, Aaron Maschinot, Jos{\'{e}} Lezama, Lu Jiang, Ming{-}Hsuan Yang, Kevin~Patrick Murphy, William~T. Freeman, Michael Rubinstein, Yuanzhen Li, and Dilip Krishnan.
\newblock Muse: Text-to-image generation via masked generative transformers.
\newblock In \emph{International Conference on Machine Learning, {ICML} 2023, 23-29 July 2023, Honolulu, Hawaii, {USA}}, pages 4055--4075. {PMLR}, 2023.

\bibitem[Chen et~al.(2021)Chen, Li, and Xue]{chen2021scene}
Jingye Chen, Bin Li, and Xiangyang Xue.
\newblock Scene text telescope: Text-focused scene image super-resolution.
\newblock In \emph{CVPR}, pages 12026--12035, 2021.

\bibitem[Chen et~al.(2023)Chen, Wang, Zhou, Qiao, and Dong]{chen2023activating}
Xiangyu Chen, Xintao Wang, Jiantao Zhou, Yu Qiao, and Chao Dong.
\newblock Activating more pixels in image super-resolution transformer.
\newblock In \emph{CVPR}, pages 22367--22377, 2023.

\bibitem[Cheng et~al.(2025)Cheng, Yu, Tu, He, Chen, Guo, Zhu, Wang, Gao, and Hu]{cheng2025effective}
Kun Cheng, Lei Yu, Zhijun Tu, Xiao He, Liyu Chen, Yong Guo, Mingrui Zhu, Nannan Wang, Xinbo Gao, and Jie Hu.
\newblock Effective diffusion transformer architecture for image super-resolution.
\newblock In \emph{Proceedings of the AAAI Conference on Artificial Intelligence}, pages 2455--2463, 2025.

\bibitem[Cui et~al.(2025)Cui, Sun, Lin, Gao, Zhang, Liu, Wang, Zhang, Zhou, Liu, Zhang, Lv, Huang, Zhang, Zhang, Zhang, Liu, Yu, and Ma]{cui2025paddleocr30technicalreport}
Cheng Cui, Ting Sun, Manhui Lin, Tingquan Gao, Yubo Zhang, Jiaxuan Liu, Xueqing Wang, Zelun Zhang, Changda Zhou, Hongen Liu, Yue Zhang, Wenyu Lv, Kui Huang, Yichao Zhang, Jing Zhang, Jun Zhang, Yi Liu, Dianhai Yu, and Yanjun Ma.
\newblock Paddleocr 3.0 technical report, 2025.

\bibitem[Deshmukh et~al.(2024)Deshmukh, Susladkar, Makwana, Mittal, et~al.]{deshmukh2024textual}
Gayatri Deshmukh, Onkar Susladkar, Dhruv Makwana, Sparsh Mittal, et~al.
\newblock Textual alchemy: Coformer for scene text understanding.
\newblock In \emph{Proceedings of the IEEE/CVF Winter Conference on Applications of Computer Vision}, pages 2931--2941, 2024.

\bibitem[Dhariwal and Nichol(2021)]{dhariwal2021diffusion}
Prafulla Dhariwal and Alexander Nichol.
\newblock Diffusion models beat gans on image synthesis.
\newblock \emph{Advances in neural information processing systems}, 34:\penalty0 8780--8794, 2021.

\bibitem[Ding et~al.(2020)Ding, Ma, Wang, and Simoncelli]{ding2020image}
Keyan Ding, Kede Ma, Shiqi Wang, and Eero~P Simoncelli.
\newblock Image quality assessment: Unifying structure and texture similarity.
\newblock \emph{TPAMI}, 44\penalty0 (5):\penalty0 2567--2581, 2020.

\bibitem[Dong et~al.(2014)Dong, Loy, He, and Tang]{dong2014learning}
Chao Dong, Chen~Change Loy, Kaiming He, and Xiaoou Tang.
\newblock Learning a deep convolutional network for image super-resolution.
\newblock In \emph{ECCV}, pages 184--199, 2014.

\bibitem[Dong et~al.(2015)Dong, Zhu, Deng, Loy, and Qiao]{dong2015boosting}
Chao Dong, Ximei Zhu, Yubin Deng, Chen~Change Loy, and Yu Qiao.
\newblock Boosting optical character recognition: A super-resolution approach.
\newblock \emph{arXiv preprint}, 2015.

\bibitem[Dong et~al.(2025{\natexlab{a}})Dong, Fan, Guo, Wang, Zhang, Chen, Luo, and Zou]{DBLP:conf/cvpr/DongFGWZCLZ25}
Linwei Dong, Qingnan Fan, Yihong Guo, Zhonghao Wang, Qi Zhang, Jinwei Chen, Yawei Luo, and Changqing Zou.
\newblock {TSD-SR:} one-step diffusion with target score distillation for real-world image super-resolution.
\newblock In \emph{{IEEE/CVF} Conference on Computer Vision and Pattern Recognition, {CVPR} 2025, Nashville, TN, USA, June 11-15, 2025}, pages 23174--23184. Computer Vision Foundation / {IEEE}, 2025{\natexlab{a}}.

\bibitem[Dong et~al.(2025{\natexlab{b}})Dong, Fan, Guo, Wang, Zhang, Chen, Luo, and Zou]{dong2025tsd}
Linwei Dong, Qingnan Fan, Yihong Guo, Zhonghao Wang, Qi Zhang, Jinwei Chen, Yawei Luo, and Changqing Zou.
\newblock Tsd-sr: One-step diffusion with target score distillation for real-world image super-resolution.
\newblock In \emph{Proceedings of the Computer Vision and Pattern Recognition Conference}, pages 23174--23184, 2025{\natexlab{b}}.

\bibitem[Duan et~al.(2025)Duan, Zhang, Jin, Zhang, Xiong, Zou, Ren, Guo, and Li]{duan2025dit4sr}
Zheng-Peng Duan, Jiawei Zhang, Xin Jin, Ziheng Zhang, Zheng Xiong, Dongqing Zou, Jimmy Ren, Chun-Le Guo, and Chongyi Li.
\newblock Dit4sr: Taming diffusion transformer for real-world image super-resolution.
\newblock In \emph{Proceedings of the IEEE/CVF International Conference on Computer Vision}, 2025.

\bibitem[Esser et~al.(2024)Esser, Kulal, Blattmann, Entezari, M{\"u}ller, Saini, Levi, Lorenz, Sauer, Boesel, et~al.]{esser2024scaling}
Patrick Esser, Sumith Kulal, Andreas Blattmann, Rahim Entezari, Jonas M{\"u}ller, Harry Saini, Yam Levi, Dominik Lorenz, Axel Sauer, Frederic Boesel, et~al.
\newblock Scaling rectified flow transformers for high-resolution image synthesis.
\newblock In \emph{Forty-first international conference on machine learning}, 2024.

\bibitem[Heusel et~al.(2017)Heusel, Ramsauer, Unterthiner, Nessler, and Hochreiter]{heusel2017gans}
Martin Heusel, Hubert Ramsauer, Thomas Unterthiner, Bernhard Nessler, and Sepp Hochreiter.
\newblock Gans trained by a two time-scale update rule converge to a local nash equilibrium.
\newblock \emph{NeurIPS}, 30, 2017.

\bibitem[Ho et~al.(2020)Ho, Jain, and Abbeel]{ho2020denoising}
Jonathan Ho, Ajay Jain, and Pieter Abbeel.
\newblock Denoising diffusion probabilistic models.
\newblock \emph{Advances in Neural Information Processing Systems}, 33:\penalty0 6840--6851, 2020.

\bibitem[Hu et~al.(2025)Hu, Fan, Luo, Yu, Guo, and Fan]{hu2025text}
Qiming Hu, Linlong Fan, Yiyan Luo, Yuhang Yu, Xiaojie Guo, and Qingnan Fan.
\newblock Text-aware real-world image super-resolution via diffusion model with joint segmentation decoders.
\newblock \emph{arXiv preprint arXiv:2506.04641}, 2025.

\bibitem[Kil et~al.(2023)Kil, Changpinyo, Chen, Hu, Goodman, Chao, and Soricut]{kil2023prestu}
Jihyung Kil, Soravit Changpinyo, Xi Chen, Hexiang Hu, Sebastian Goodman, Wei-Lun Chao, and Radu Soricut.
\newblock Prestu: Pre-training for scene-text understanding.
\newblock In \emph{Proceedings of the IEEE/CVF international conference on computer vision}, pages 15270--15280, 2023.

\bibitem[Kingma and Welling(2013)]{kingma2013auto}
Diederik~P Kingma and Max Welling.
\newblock Auto-encoding variational bayes.
\newblock \emph{arXiv preprint arXiv:1312.6114}, 2013.

\bibitem[Künzel et~al.(2025)Künzel, Hilsmann, and Eisert]{ripe2025}
Johannes Künzel, Anna Hilsmann, and Peter Eisert.
\newblock {RIPE: Reinforcement Learning on Unlabeled Image Pairs for Robust Keypoint Extraction}.
\newblock \emph{arXiv}, 2025.

\bibitem[Li et~al.(2022)Li, Liu, Guo, Yin, Jiang, Du, Du, Zhu, Lai, Hu, Yu, and Ma]{DBLP:journals/corr/abs-2206-03001}
Chenxia Li, Weiwei Liu, Ruoyu Guo, Xiaoting Yin, Kaitao Jiang, Yongkun Du, Yuning Du, Lingfeng Zhu, Baohua Lai, Xiaoguang Hu, Dianhai Yu, and Yanjun Ma.
\newblock Pp-ocrv3: More attempts for the improvement of ultra lightweight {OCR} system.
\newblock \emph{CoRR}, abs/2206.03001, 2022.

\bibitem[Li and Cui(2025)]{DBLP:conf/iclr/LiC25}
Peidong Li and Dixiao Cui.
\newblock Navigation-guided sparse scene representation for end-to-end autonomous driving.
\newblock In \emph{The Thirteenth International Conference on Learning Representations, {ICLR} 2025, Singapore, April 24-28, 2025}. OpenReview.net, 2025.

\bibitem[Li et~al.(2023{\natexlab{a}})Li, Zuo, and Loy]{li2023learning}
Xiaoming Li, Wangmeng Zuo, and Chen~Change Loy.
\newblock Learning generative structure prior for blind text image super-resolution.
\newblock In \emph{CVPR}, pages 10103--10113, 2023{\natexlab{a}}.

\bibitem[Li et~al.(2023{\natexlab{b}})Li, Zhang, Liang, Cao, Liu, Gong, Zhang, Tang, Liu, Demandolx, et~al.]{li2023lsdir}
Yawei Li, Kai Zhang, Jingyun Liang, Jiezhang Cao, Ce Liu, Rui Gong, Yulun Zhang, Hao Tang, Yun Liu, Denis Demandolx, et~al.
\newblock Lsdir: A large scale dataset for image restoration.
\newblock In \emph{CVPRW}, pages 1775--1787, 2023{\natexlab{b}}.

\bibitem[Lin et~al.(2024)Lin, He, Chen, Lyu, Dai, Yu, Qiao, Ouyang, and Dong]{lin2024diffbir}
Xinqi Lin, Jingwen He, Ziyan Chen, Zhaoyang Lyu, Bo Dai, Fanghua Yu, Yu Qiao, Wanli Ouyang, and Chao Dong.
\newblock Diffbir: Toward blind image restoration with generative diffusion prior.
\newblock In \emph{ECCV}, pages 430--448, 2024.

\bibitem[Loshchilov and Hutter(2017)]{loshchilov2017decoupled}
Ilya Loshchilov and Frank Hutter.
\newblock Decoupled weight decay regularization.
\newblock \emph{arXiv preprint arXiv:1711.05101}, 2017.

\bibitem[Lowe(1999)]{lowe1999object}
David~G Lowe.
\newblock Object recognition from local scale-invariant features.
\newblock In \emph{Proceedings of the seventh IEEE international conference on computer vision}, pages 1150--1157. Ieee, 1999.

\bibitem[Ma et~al.(2022)Ma, Liang, and Zhang]{ma2022text}
Jianqi Ma, Zhetong Liang, and Lei Zhang.
\newblock A text attention network for spatial deformation robust scene text image super-resolution.
\newblock In \emph{CVPR}, pages 5911--5920, 2022.

\bibitem[Ma et~al.(2023)Ma, Liang, Xiang, Yang, and Zhang]{ma2023benchmark}
Jianqi Ma, Zhetong Liang, Wangmeng Xiang, Xi Yang, and Lei Zhang.
\newblock A benchmark for chinese-english scene text image super-resolution.
\newblock In \emph{ICCV}, pages 19452--19461, 2023.

\bibitem[Mou et~al.(2020)Mou, Tan, Yang, Chen, Liu, Yan, and Huang]{mou2020plugnet}
Yongqiang Mou, Lei Tan, Hui Yang, Jingying Chen, Leyuan Liu, Rui Yan, and Yaohong Huang.
\newblock Plugnet: Degradation aware scene text recognition supervised by a pluggable super-resolution unit.
\newblock In \emph{ECCV}, pages 158--174, 2020.

\bibitem[Ramesh et~al.(2022)Ramesh, Dhariwal, Nichol, Chu, and Chen]{ramesh2022hierarchical}
Aditya Ramesh, Prafulla Dhariwal, Alex Nichol, Casey Chu, and Mark Chen.
\newblock Hierarchical text-conditional image generation with clip latents.
\newblock \emph{arXiv preprint arXiv:2204.06125}, 2022.

\bibitem[Roberts and Mullis(1987)]{roberts1987digital}
Richard~A Roberts and Clifford~T Mullis.
\newblock \emph{Digital signal processing}.
\newblock Addison-Wesley Longman Publishing Co., Inc., 1987.

\bibitem[Rombach et~al.(2022)Rombach, Blattmann, Lorenz, Esser, and Ommer]{rombach2022high}
Robin Rombach, Andreas Blattmann, Dominik Lorenz, Patrick Esser, and Bj{\"o}rn Ommer.
\newblock High-resolution image synthesis with latent diffusion models.
\newblock In \emph{CVPR}, pages 10684--10695, 2022.

\bibitem[Saharia et~al.(2022)Saharia, Chan, Saxena, Li, Whang, Denton, Ghasemipour, Gontijo~Lopes, Karagol~Ayan, Salimans, et~al.]{saharia2022photorealistic}
Chitwan Saharia, William Chan, Saurabh Saxena, Lala Li, Jay Whang, Emily~L Denton, Kamyar Ghasemipour, Raphael Gontijo~Lopes, Burcu Karagol~Ayan, Tim Salimans, et~al.
\newblock Photorealistic text-to-image diffusion models with deep language understanding.
\newblock \emph{Advances in neural information processing systems}, 35:\penalty0 36479--36494, 2022.

\bibitem[Souibgui et~al.(2023)Souibgui, Biswas, Mafla, Biten, Forn{\'e}s, Kessentini, Llad{\'o}s, Gomez, and Karatzas]{souibgui2023text}
Mohamed~Ali Souibgui, Sanket Biswas, Andres Mafla, Ali~Furkan Biten, Alicia Forn{\'e}s, Yousri Kessentini, Josep Llad{\'o}s, Lluis Gomez, and Dimosthenis Karatzas.
\newblock Text-diae: a self-supervised degradation invariant autoencoder for text recognition and document enhancement.
\newblock In \emph{proceedings of the AAAI conference on artificial intelligence}, pages 2330--2338, 2023.

\bibitem[Tang et~al.(2023)Tang, Miyazaki, and Omachi]{tang2023scene}
Zhengmi Tang, Tomo Miyazaki, and Shinichiro Omachi.
\newblock A scene-text synthesis engine achieved through learning from decomposed real-world data.
\newblock \emph{IEEE Transactions on Image Processing}, 32:\penalty0 5837--5851, 2023.

\bibitem[Tuo et~al.(2024)Tuo, Xiang, He, Geng, and Xie]{DBLP:conf/iclr/TuoXHGX24}
Yuxiang Tuo, Wangmeng Xiang, Jun{-}Yan He, Yifeng Geng, and Xuansong Xie.
\newblock Anytext: Multilingual visual text generation and editing.
\newblock In \emph{The Twelfth International Conference on Learning Representations, {ICLR} 2024, Vienna, Austria, May 7-11, 2024}. OpenReview.net, 2024.

\bibitem[Wang et~al.(2024{\natexlab{a}})Wang, Shen, and Zhang]{wang2024multi}
Bolin Wang, Xinyi Shen, and Kejun Zhang.
\newblock Multi-task diffusion model for simultaneous text and image inpainting.
\newblock In \emph{2024 17th International Symposium on Computational Intelligence and Design (ISCID)}, pages 301--305. IEEE, 2024{\natexlab{a}}.

\bibitem[Wang et~al.(2024{\natexlab{b}})Wang, Yue, Zhou, Chan, and Loy]{wang2024exploiting}
Jianyi Wang, Zongsheng Yue, Shangchen Zhou, Kelvin~CK Chan, and Chen~Change Loy.
\newblock Exploiting diffusion prior for real-world image super-resolution.
\newblock \emph{IJCV}, 132\penalty0 (12):\penalty0 5929--5949, 2024{\natexlab{b}}.

\bibitem[Wang et~al.(2019)Wang, Xie, Sun, Wang, Tian, Shen, and Luo]{wang2019textsr}
Wenjia Wang, Enze Xie, Peize Sun, Wenhai Wang, Lixun Tian, Chunhua Shen, and Ping Luo.
\newblock Textsr: Content-aware text super-resolution guided by recognition.
\newblock \emph{arXiv preprint}, 2019.

\bibitem[Wang et~al.(2020)Wang, Xie, Liu, Wang, Liang, Shen, and Bai]{wang2020scene}
Wenjia Wang, Enze Xie, Xuebo Liu, Wenhai Wang, Ding Liang, Chunhua Shen, and Xiang Bai.
\newblock Scene text image super-resolution in the wild.
\newblock In \emph{ECCV}, pages 650--666, 2020.

\bibitem[Wang et~al.(2021)Wang, Xie, Dong, and Shan]{wang2021real}
Xintao Wang, Liangbin Xie, Chao Dong, and Ying Shan.
\newblock Real-esrgan: Training real-world blind super-resolution with pure synthetic data.
\newblock In \emph{ICCV}, pages 1905--1914, 2021.

\bibitem[Wang et~al.(2024{\natexlab{c}})Wang, Yang, Chen, Wang, Guo, Chau, Liu, Qiao, Kot, and Wen]{wang2024sinsr}
Yufei Wang, Wenhan Yang, Xinyuan Chen, Yaohui Wang, Lanqing Guo, Lap-Pui Chau, Ziwei Liu, Yu Qiao, Alex~C Kot, and Bihan Wen.
\newblock Sinsr: diffusion-based image super-resolution in a single step.
\newblock In \emph{CVPR}, pages 25796--25805, 2024{\natexlab{c}}.

\bibitem[Wu et~al.(2024{\natexlab{a}})Wu, Sun, Ma, and Zhang]{wu2025one}
Rongyuan Wu, Lingchen Sun, Zhiyuan Ma, and Lei Zhang.
\newblock One-step effective diffusion network for real-world image super-resolution.
\newblock \emph{NeurIPS}, 37:\penalty0 92529--92553, 2024{\natexlab{a}}.

\bibitem[Wu et~al.(2024{\natexlab{b}})Wu, Yang, Sun, Zhang, Li, and Zhang]{wu2024seesr}
Rongyuan Wu, Tao Yang, Lingchen Sun, Zhengqiang Zhang, Shuai Li, and Lei Zhang.
\newblock Seesr: Towards semantics-aware real-world image super-resolution.
\newblock In \emph{CVPR}, pages 25456--25467, 2024{\natexlab{b}}.

\bibitem[Yang et~al.(2024)Yang, Wu, Ren, Xie, and Zhang]{yang2024pixel}
Tao Yang, Rongyuan Wu, Peiran Ren, Xuansong Xie, and Lei Zhang.
\newblock Pixel-aware stable diffusion for realistic image super-resolution and personalized stylization.
\newblock In \emph{ECCV}, pages 74--91, 2024.

\bibitem[Ye et~al.(2024)Ye, Zhang, Liu, Liu, Yin, Liu, Du, and Tao]{ye2024hi}
Maoyuan Ye, Jing Zhang, Juhua Liu, Chenyu Liu, Baocai Yin, Cong Liu, Bo Du, and Dacheng Tao.
\newblock Hi-sam: Marrying segment anything model for hierarchical text segmentation.
\newblock \emph{TPAMI}, 2024.

\bibitem[Yu et~al.(2024)Yu, Gu, Li, Hu, Kong, Wang, He, Qiao, and Dong]{yu2024scaling}
Fanghua Yu, Jinjin Gu, Zheyuan Li, Jinfan Hu, Xiangtao Kong, Xintao Wang, Jingwen He, Yu Qiao, and Chao Dong.
\newblock Scaling up to excellence: Practicing model scaling for photo-realistic image restoration in the wild.
\newblock In \emph{CVPR}, pages 25669--25680, 2024.

\bibitem[Yu et~al.(2021)Yu, Chen, Li, Ma, Guan, Xu, Wang, Qu, and Xue]{yu2021benchmarking}
Haiyang Yu, Jingye Chen, Bin Li, Jianqi Ma, Mengnan Guan, Xixi Xu, Xiaocong Wang, Shaobo Qu, and Xiangyang Xue.
\newblock Benchmarking chinese text recognition: Datasets, baselines, and an empirical study.
\newblock \emph{arXiv preprint arXiv:2112.15093}, 2021.

\bibitem[Yu et~al.(2023)Yu, Wang, Li, and Xue]{yu2023chinese}
Haiyang Yu, Xiaocong Wang, Bin Li, and Xiangyang Xue.
\newblock Chinese text recognition with a pre-trained clip-like model through image-ids aligning.
\newblock In \emph{Proceedings of the IEEE/CVF International Conference on Computer Vision}, pages 11943--11952, 2023.

\bibitem[Yue et~al.(2023)Yue, Wang, and Loy]{yue2023resshift}
Zongsheng Yue, Jianyi Wang, and Chen~Change Loy.
\newblock Resshift: Efficient diffusion model for image super-resolution by residual shifting.
\newblock \emph{NeurIPS}, 36:\penalty0 13294--13307, 2023.

\bibitem[Yujian and Bo(2007)]{yujian2007normalized}
Li Yujian and Liu Bo.
\newblock A normalized levenshtein distance metric.
\newblock \emph{TPAMI}, 29\penalty0 (6):\penalty0 1091--1095, 2007.

\bibitem[Zhang et~al.(2021)Zhang, Liang, Van~Gool, and Timofte]{zhang2021designing}
Kai Zhang, Jingyun Liang, Luc Van~Gool, and Radu Timofte.
\newblock Designing a practical degradation model for deep blind image super-resolution.
\newblock In \emph{ICCV}, pages 4791--4800, 2021.

\bibitem[Zhang et~al.(2023)Zhang, Rao, and Agrawala]{zhang2023adding}
Lvmin Zhang, Anyi Rao, and Maneesh Agrawala.
\newblock Adding conditional control to text-to-image diffusion models.
\newblock In \emph{ICCV}, pages 3836--3847, 2023.

\bibitem[Zhang et~al.(2018)Zhang, Isola, Efros, Shechtman, and Wang]{zhang2018unreasonable}
Richard Zhang, Phillip Isola, Alexei~A Efros, Eli Shechtman, and Oliver Wang.
\newblock The unreasonable effectiveness of deep features as a perceptual metric.
\newblock In \emph{CVPR}, pages 586--595, 2018.

\bibitem[Zhang et~al.(2024{\natexlab{a}})Zhang, Zhang, Li, Wang, Hou, Zou, and Bian]{zhang2024diffusion}
Yuzhe Zhang, Jiawei Zhang, Hao Li, Zhouxia Wang, Luwei Hou, Dongqing Zou, and Liheng Bian.
\newblock Diffusion-based blind text image super-resolution.
\newblock In \emph{CVPR}, pages 25827--25836, 2024{\natexlab{a}}.

\bibitem[Zhang et~al.(2024{\natexlab{b}})Zhang, Zhang, Xing, Li, Zhao, Sun, Lan, Luan, Huang, and Lin]{zhang2024artbank}
Zhanjie Zhang, Quanwei Zhang, Wei Xing, Guangyuan Li, Lei Zhao, Jiakai Sun, Zehua Lan, Junsheng Luan, Yiling Huang, and Huaizhong Lin.
\newblock Artbank: Artistic style transfer with pre-trained diffusion model and implicit style prompt bank.
\newblock In \emph{Proceedings of the AAAI conference on artificial intelligence}, pages 7396--7404, 2024{\natexlab{b}}.

\end{thebibliography}
}

\clearpage
\setcounter{page}{1}
\maketitlesupplementary
\begin{table*}[t]
\caption{Statistics of dataset size and line count in subsets of UltraZoom-ST.}
\vspace{-15pt}
\label{tab:dataset_detail}
\begin{center}
\begin{tabular}{lcccc} 
\toprule
\bf Subset
& \bf image count
& \bf line count
& \bf mean lines/img 
& \bf img $>$ 5 lines

\\
\midrule
14 mm
& 1,439
& 15,073
& 10.47
& 754
\\
35 mm 
& 1,798
& 17,263
& 9.60
& 867
\\
85 mm
&  1,799
& 17,339
& 9.64
& 869
\\

\bottomrule

\end{tabular}
\end{center}
\end{table*}
\begin{table*}[t]
\caption{Alignment comparison.
}
\vspace{-15pt}
\label{tab:align}
\begin{center}
\begin{tabular}{lccccc} 
\toprule
Alignment Method
& PSNR $\uparrow$
& MSE $\downarrow$
& SSIM $\uparrow$
& NCC $\uparrow$
& AKD $\downarrow$
\\
\midrule
RealSR
& 13.87
& 4002.65
& 0.5072
& 0.3285
& 546.78
\\
SIFT Alignment (Single time)
& 12.77
& 6332.79
& 0.4896
& 0.4313
& 546.78
\\
\rowcolor{gray!30} \bf Cascade Coarse-to-Fine Alignment (Ours)
& \bf 23.00
& \bf 441.52
& \bf 0.7279
& \bf 0.9184
& \bf 241.44
\\
\bottomrule

\end{tabular}
\end{center}
\end{table*}
\begin{table*}[t]
\caption{Efficiency analysis.}
\vspace{-15pt}
\label{tab:Efficiency}
\begin{center}
\begin{tabular}{lccc} 
\toprule
\bf Methods
& \bf Flops (GFLOPs)
& \bf Speed (ms)
& \bf OCR-A
\\
\midrule
HAT
& 6670.32
& 1086.68
& 37.9\%
\\
DiffTSR
& 58502.14
& 8610.59
& 31.7\%
\\
DiT4SR w/o llava
& 160787.14
& 17385.14
& 23.7\%
\\
DreamClear w/o llava
& 412843.67
& 83193.81
& 21.4\%
\\
TADiSR
& 4497.96
& 342.32
& 36.6\%
\\
TiGeSR (ours) Stage 1
& 6215.85
& 663.5
& \multirow{2}{*}{43.0\%}
\\
TiGeSR (ours) Stage 2
& 3734.01
& 360.06
& ~
\\
\bottomrule
\end{tabular}
\end{center}
\end{table*}

\begin{table*}[t]
\caption{Ablation on the effect of stochastic OCR outputs on the performance of our method. 
The performance gains as the randomness of the OCR output drops. 
Even under 100\% random OCR output, our method still outperforms TADiSR, proving low reliance on OCR.  }
\label{tab:ocr_ablation_detail}
\begin{center}
\vspace{-15pt}
\begin{tabular}{lcccccc} 
\toprule
\bf Methods
& \bf PSNR$\uparrow$
& \bf SSIM$\uparrow$
& \bf LPIPS$\downarrow$
& \bf DISTS$\downarrow$
& \bf FID$\downarrow$
& \bf OCR-A$\uparrow$
\\ 

\midrule
TADiSR
& 24.68
& 0.795
& 0.362
& 0.227
& 52.13
& 35.5\%
\\
\midrule
\multicolumn{7}{c}{TiGeSR (Ours) performance under different randomness of OCR output.}
\\
100\% Random Text
& 25.60
& 0.833
& 0.188
& \textbf{0.150}
& 28.92
& 40.3\%
\\
50\% Random Text
& 25.60
& 0.833
& 0.189
& 0.151
& 28.92
& 40.6\%
\\
20\% Random Text
& \textbf{25.63}
& \textbf{0.834}
& \textbf{0.185}
& \textbf{0.150}
& 29.3
& 43.7\%
\\
\textbf{0\% Random Text (Ours)}
& \textbf{25.63}
& \textbf{0.834}
& \textbf{0.185}
& \textbf{0.150}
& \textbf{28.82}
& \textbf{44.6}\%
\\
\bottomrule
\end{tabular}
\end{center}
\end{table*}

\begin{table*}[t]
\footnotesize
\caption{Evaluation Results on Both Image Quality and Text Accuracy on UltraZoom-ST.}
\vspace{-15pt}
\label{tab:Ultrazoom}
\begin{center}
\begin{tabular}{clcccccccccc}  
\toprule
~ 
&  \multicolumn{1}{l}{\bf Methods} 
&  \multicolumn{1}{c}{\bf PSNR$\uparrow$}  
&  \multicolumn{1}{c}{\bf SSIM$\uparrow$ }
&  \multicolumn{1}{c}{\bf LPIPS$\downarrow$ }
&  \multicolumn{1}{c}{\bf DISTS$\downarrow$ }
&  \multicolumn{1}{c}{\bf FID$\downarrow$ }
&  \multicolumn{1}{c}{\bf PSNR$_{cr}$$\uparrow$ }
&  \multicolumn{1}{c}{\bf SSIM$_{cr}$$\uparrow$ }
&  \multicolumn{1}{c}{\bf LPIPS$_{cr}$$\downarrow$ }
&  \multicolumn{1}{c}{\bf DISTS$_{cr}$$\downarrow$ }
&  \multicolumn{1}{c}{\bf OCR-A$\uparrow$}
\\ 
\midrule
\multirow{13}{*}{\rotatebox{90}{85mm ($\times 2.35$)}} & Real-ESRGAN 
& 25.84
& 0.854
& 0.139
& \uline{0.130}
& 28.25
& 23.59
& 0.843
& 0.170
& 0.168
& 60.8\%

  \\
~ &
HAT 
& \textbf{27.33}
& \textbf{0.873}
& \uline{0.128}
& 0.133
& 27.21
& \uline{24.76}
& \bf{0.870}
& \uline{0.139}
& \uline{0.146}
& \uline{62.0\%}
  \\
~ &
MACRONet 
& 22.90
& 0.804
& 0.234
& 0.173
& 40.19
& 17.92
& 0.693
& 0.515
& 0.360
& 54.2\%
  \\
~ &
SeeSR 
& 25.02	
& 0.826
& 0.167
& 0.159
& 32.31
& 21.62
& 0.796
& 0.245
& 0.207
& 28.2\%

  \\
~ &
SupIR 
& 25.58
& 0.823
& 0.202
& 0.160
& 29.33
& 21.91
& 0.808
& 0.239
& 0.214
& 33.3\%
 
  \\
~ &
DiffTSR 
& 23.48
& 0.805
& 0.230
& 0.157
& 35.77
& 17.92
& 0.688
& 0.392
& 0.285
& 48.2\%

  \\
~ &
DiffBIR 
& 25.25	 
& 0.793
& 0.175
& 0.152
& 24.57
& 22.26
& 0.803
& 0.218
& 0.190
& 36.8\%
\\
~ &
OSEDiff 
& 26.38
& 0.852
& 0.153
& 0.137
& 23.92
& 22.84
& 0.833
& 0.200
& 0.204
& 41.0\%

\\
~ &
DreamClear 
& 25.31
& 0.811
& 0.169
& 0.151
& \uline{24.19}
& 21.44
& 0.790
& 0.253
& 0.221
& 26.8\%
  \\
~ &
TSD-SR 
& 23.99
& 0.800
& 0.161
& 0.165
& 30.40
& 20.62
& 0.786
& 0.232
& 0.218
& 37.8\%
\\
~ &
DiT4SR 
& 24.71
& 0.810
& 0.173
& 0.143
& 29.15	
& 21.25
& 0.800
& 0.221
& 0.193
& 48.2\%
\\
~ &
TADiSR 
& 26.31
& 0.858
& 0.177
& 0.149
& 30.47
& 23.55
& 0.856
& 0.175
& 0.178
& 55.7\%
\\
~ 
& Ours 
& \uline{27.00}
& \uline{0.871}
& \textbf{0.123}
& \textbf{0.120}
& \textbf{22.34}
& \textbf{24.31}
& \uline{0.860}
& \textbf{0.135}
& \textbf{0.142}
& \textbf{63.2\%}

  \\ 
\midrule
\midrule
\multirow{13}{*}{\rotatebox{90}{35mm ($\times 5.71$)}} 
& Real-ESRGAN
& 23.70
& 0.785
& 0.218
& 0.184
& 42.62
& 20.86
& 0.782
& 0.272
& 0.235
& 34.1\%

  \\
~ 
& HAT 
& 25.05
& 0.819
& 0.207
& 0.182
& 41.32
& \uline{21.98}
& \uline{0.816}
& \uline{0.250}
& 0.241
& 34.6\%

  \\
~ 
& MACRONet
& 22.28
& 0.774
& 0.284
& 0.199
& 49.14
& 17.43
& 0.694
& 0.523
& 0.361
& 31.5\%
 
  \\
  ~ &
SeeSR 
& 23.76
& 0.795 
& 0.198
& 0.175
& 37.92
& 20.55
& 0.778	
& 0.271	
& 0.222
& 25.2\%
  \\
  ~ &
SupIR 
& 23.49
& 0.746
& 0.302
& 0.204
& 41.80
& 20.00	
& 0.757	
& 0.335	
& 0.282
& 24.0\%
  \\
  ~ &
DiffTSR 
& 22.57
& 0.773
& 0.281
& 0.184
& 45.78
& 16.99	
& 0.671	
& 0.443	
& 0.312
& 31.2\%

  \\
  ~ &
DiffBIR 
& 23.76
& 0.731
& 0.245
& 0.196
& 35.99	
& 20.82
& 0.770	
& 0.275	
& 0.234
& 28.6\%
  \\
  ~ &
OSEDiff
  & \uline{25.35}
  & \uline{0.827}
  & \uline{0.192}
  & 0.166
  & \uline{29.06}
  & 21.62
  & 0.809	
  & 0.263	
  & 0.263
  & 29.9\%
  \\
  ~ &
DreamClear 
& 24.20
& 0.778
& 0.209
& 0.177
& 32.08
& 20.66	
& 0.779	
& 0.283	
& 0.251
& 23.9\%
  \\
  ~ &
TSD-SR 
& 22.93
& 0.757
& 0.200
& 0.199
& 37.61
& 19.64	
& 0.760	
& 0.282	
& 0.246
& 27.3\%
  \\
  ~ &
DiT4SR 
& 23.45
& 0.774
& 0.204
& \uline{0.158}
& 31.73
& 20.12	
& 0.779	
& \uline{0.250}
& \uline{0.208}
& 26.5\%
  \\
~ 
& TADiSR 
& 24.68
& 0.795
& 0.362
& 0.227
& 52.13
& 21.57
& 0.798
& 0.366	
& 0.347
& \uline{35.5\%}
  \\
~ 
& Ours 
& \textbf{25.63}
& \textbf{0.834}
& \textbf{0.185}
& \textbf{0.150}
& \textbf{28.82}
& \textbf{22.14}	
& \textbf{0.823}
& \textbf{0.204}	
& \textbf{0.194}
& \textbf{44.6\%}
\\
\midrule
\midrule
\multirow{13}{*}{\rotatebox{90}{14mm ($\times 14.29$)}} & Real-ESRGAN
& 22.09
& 0.718
& 0.423
& 0.284
& 89.22
& 18.87	
& 0.721	
& 0.503	
& 0.427
& 11.5\%
  \\
  ~ &
HAT 
& 22.66
& 0.738
& 0.451
& 0.299
& 89.58
& \uline{19.27}	
& \uline{0.738}	
& 0.533	
& 0.484
& 12.2\%
  \\
  ~ &
MACRONet
& 21.00
& 0.716
& 0.422
& 0.254
& 79.67
& 17.01
& 0.680	
& 0.553	
& 0.382
& 10.1\%
  \\
  ~ &
SeeSR 
& 21.79
& 0.731
& 0.310
& 0.228
& 64.31
& 18.87	
& 0.717	
& 0.396	
& 0.308
& 13.2\%
  \\
  ~ &
SupIR 
& 21.38
& 0.681
& 0.447
& 0.268
& 70.09
& 17.99	
& 0.690	
& 0.487	
& 0.383
& 11.0\%
  \\
~ &
DiffTSR 
& 20.88
& 0.713
& 0.412
& 0.233
& 74.07
& 15.87	
& 0.630	
& 0.538	
& 0.348
& 12.0\%
  \\
~ &
DiffBIR 
& 21.61
& 0.630
& 0.392
& 0.255
& 69.64
& 18.61	
& 0.685	
& 0.438	
& 0.328
& 11.3\%
  \\
~ &
OSEDiff
& \uline{23.11}
& \uline{0.768}
& \uline{0.274}
& 0.214
& 49.76
& 19.44
& 0.740
& 0.385
& 0.334
& 12.7\%
\\
~ 
& DreamClear 
& 22.45
& 0.719
& 0.364
& 0.260
& 62.95
& 19.14
& 0.725
& 0.440
& 0.377
& 11.7\%

  \\
~ 
& TSD-SR 
& 21.14
& 0.705
& \textbf{0.273}
& 0.225
& 57.07
& 17.67
& 0.686
& 0.393
& \uline{0.293}
& 11.9\%
\\
~ 
& DiT4SR 
& 20.88
& 0.707
& 0.282
& \textbf{0.181}
& \textbf{47.45}
& 17.36
& 0.687	
& \textbf{0.367}
& \textbf{0.253}
& 11.7\%
  \\
~ 
& TADiSR 
& 22.43
& 0.721
& 0.555
& 0.320
& 98.44
& 19.20
& 0.730
& 0.585
& 0.518
& \uline{14.4\%}
  \\
~ 
& Ours 
& \textbf{23.41}
& \textbf{0.774}
& 0.301	
& \uline{0.209}
& \uline{54.74}
& \textbf{19.73}	
& \textbf{0.748}
& \uline{0.374}
& 0.322
& \textbf{16.0\%}
  \\ 

  \midrule
  \midrule
  \multirow{13}{*}{\rotatebox{90}{Total}} & RealEsrGAN &
  23.99 &
  0.790 &
  0.248 &
  0.194 &
  30.60
  & 21.25
  & 0.786
  & 0.302
  & 0.266
& 37.1\%
  \\
  ~ &
HAT &
  \uline{25.17} &
  0.815 &
  0.249 &
  0.198 &
  30.12
  & \uline{22.18}
& \uline{0.813}
& 0.291
& 0.276
  & 37.9\%
  \\
  ~ &
MARCONet &
  22.13 &
  0.768 &
  0.306 &
  0.205 &
  34.28
  & 17.48
& 0.690
& 0.529
& 0.366
 & 33.4\%
  \\
  ~ &
SeeSR &
  23.64 &
  0.788 &
  0.219 &
  0.184 &
  26.73
  & 20.45
& 0.767
& 0.297
& 0.241
  & 22.8\%
  \\
  ~ &
SupIR &
  23.62 &
  0.754 &
  0.308 &
  0.207 &
  27.91
  & 20.10
& 0.756
& 0.344
& 0.287
  & 23.6\%
  \\
  ~ &
DiffTSR &
  22.41 &
  0.767 &
  0.300 &
  0.189 &
  31.25
  & 17.00
& 0.665
& 0.452
& 0.313
  & 31.7\%
  \\
  ~ &
DiffBIR &
   23.67 &
   0.724 &
   0.262 &
   0.197 &
   23.10
   & 20.70
& 0.757
& 0.302
& 0.245
   & 26.6\% 
  \\
  ~ &
OSEDiff
  & 25.07
 & \uline{0.819}
 & \uline{0.201}
 & 0.169
 & \uline{20.53}
 & 21.43
& 0.798
& 0.276
& 0.263
 & 28.9\%
  \\
  ~ &
DreamClear 
& 24.10 
& 0.773 
& 0.238 
& 0.191 
& 21.75
& 20.50
& 0.768
& 0.317
& 0.276
& 21.4\%
   
  \\
  ~ &
TSD-SR &
  22.79 &
  0.757&
  0.207&
  0.194&
  24.08 
  & 19.42
& 0.748
& 0.296
& 0.249
 & 26.6\% 
  \\
  ~ &
DiT4SR &
   23.16 &
   0.767 &
   0.215 &
   \uline{0.159} &
   20.58
   & 19.73
& 0.760
& \uline{0.273}
& \uline{0.216}
   & 23.7\%
  \\
  ~ &
TADiSR &
   24.61 &
   0.796 &
   0.203 &
   0.160 &
   36.61
   & 21.59
& 0.799
& 0.360
& 0.336
   & \uline{36.6\%}  
  \\
  ~ &
 Ours &
   \textbf{25.48}  &
   \textbf{0.830} &
   \bf{0.196} &
   \bf{0.156} &
    \textbf{20.01}  
     & \textbf{22.22}
 & \textbf{0.814}
 & \textbf{0.228}
 & \textbf{0.212}
   & \textbf{43.0\%} 
  \\ 
\bottomrule

\end{tabular}
\end{center}
\end{table*}

\section{Reproducibility Statement}

To ensure reproducibility, we have made the following efforts: 
\begin{enumerate}
    \item We will release our code and dataset.
    \item We provide implementation details in \cref{sec:Imple} and \cref{Appen:Implement}, including the training process and selection of hyper-parameters.
    \item  We provide details on evaluation metrics and dataset preparation in \cref{sec:4_dataset} and \cref{Appen:dataset_collection}, and the code and data will be made available along with it.
\end{enumerate}

\section{Ethics Statement}

This work focuses on improving scene text image super-resolution to support beneficial applications such as enhancing accessibility, document restoration, and navigation assistance. 
However, we acknowledge potential risks, including misuse in privacy-sensitive contexts (e.g., recovering text from personal or social media images) or unintended deployment in surveillance. 
To mitigate such risks, users are encouraged to combine our methods with privacy-preserving techniques such as watermarking or selective inpainting. 
Our UZ-ST dataset was collected from public, non-sensitive scenes under ethical guidelines, without including personally identifiable or private information. 
We believe the societal benefits of improved text image restoration outweigh potential risks, provided that the technology is applied responsibly.

\section{LLM Acknowledgments}

We thank ChatGPT (GPT-5) and DeepSeek for the help in refining the language and improving clarity. 
All wording and factual content were reviewed and approved by the authors.

\section{Dataset Collection}
\label{Appen:dataset_collection}

We use VIVO X200 Ultra to collect images for 4 separate focal lengths (14 mm, 35 mm, 84 mm, and 200 mm).
We first use PP-OCRV5~\cite{cui2025paddleocr30technicalreport} for the rough annotation, then we manually filter the images and annotations. 
During the filtering and annotation, each image undergoes the following rules:
\begin{itemize}
    \item Width or height of the image should be no less than 256.
    \item Height of the text should not be less than 32 pixels.
    \item Score of OCR recognition of the text should not be lower than 0.9.
    \item Content of the text should not be empty or consist solely of whitespace.
\end{itemize}

\section{More Implementation Details}
\label{Appen:Implement}
\noindent \textbf{Dataset Settings.}~For training, we combine a synthetic dataset with real paired datasets (Real-CE~\citep{ma2023benchmark} and UZ-ST). 
Our synthetic dataset builds upon LSDIR~\citep{li2023lsdir}, containing 27,000 triplets $(x_H,x_L,x_m)$. 
We render text on the GT of LSDIR and the corresponding text mask using the LBTS~\citep{tang2023scene}, then apply Real-ESRGAN degradation \citet{wang2021real} to generate LR images.
Since there are misaligned image pairs in Real-CE~\citep{hu2025text,zhang2024diffusion} and text lines not annotated, we filter out misaligned image pairs and reannotate the images manually. 
In the end, we obtain 337 training pairs and 188 testing pairs from the Real-CE dataset. 
Images under 13mm and 52mm focal lengths are considered as LR $x_L$ and GT $x_H$, respectively. 
Following \citet{hu2025text}, we use SAM-TS~\citep{ye2024hi} to obtain the text mask $x_m$ from $x_H$. 
UZ-ST triplets $(x_H,x_L,x_m)$ are processed identically. 
For stage 1 training, we use cropped text regions from the dataset, while for stage 2, we use full images. 
Following~\cite{DBLP:conf/iclr/TuoXHGX24}, we use PP-OCRv3~\cite{DBLP:journals/corr/abs-2206-03001} to extract strings from the text region of predicted SR for evaluation of OCR-A.

\noindent \textbf{Training Details.}~For stage 1, we build our model based on the IDM baseline of DiffTSR~\citep{zhang2024diffusion}.
We set $\lambda_{td}$, $\lambda_{Seg}$, $\lambda_{Focal}$, and $\lambda_{Dice}$ as $1$, $0.1$, $20$, and $1$ respectively.
In stage 2, our model is built upon Stable Diffusion 3.5 medium~\citep{esser2024scaling} and follows a similar tile-based inference strategy to TADiSR~\citep{hu2025text} and SUPIR~\citep{yu2024scaling}. 
The timestep $t$ is set to $150$. 
The conditioning scale of the ControlNet is set to 1.0.
$\lambda_{l2}$, $\lambda_{LPIPS}$, and $\lambda_{edge}$ are set to $1$, $5$, $100$ when using the synthetic dataset for pretraining. 
Then, $\lambda_{edge}$ is set to $0$ and the remaining coefficients are kept unchanged when training on real paired datasets.
We train both stages of the model using the AdamW~\citep{loshchilov2017decoupled} optimizer and set the learning rate to $5 \times 10^{-5}$ and $5 \times 10^{-6}$ for stages 1 and 2 separately. 
All experiments are conducted on NVIDIA H20 GPUs. 
For stage 1, we train the model on synthetic and real datasets for 8 epochs, then finetune it on synthetic data for 2 epochs. 
For stage 2, we first pretrain the model on the synthetic dataset for 50 epochs. 
Then we train on the real paired datasets for 50 epochs. 

\section{More details of UZ-ST}
\begin{figure*}[t]
\begin{center}
\vspace{-5pt}
\includegraphics[width=0.9\linewidth]{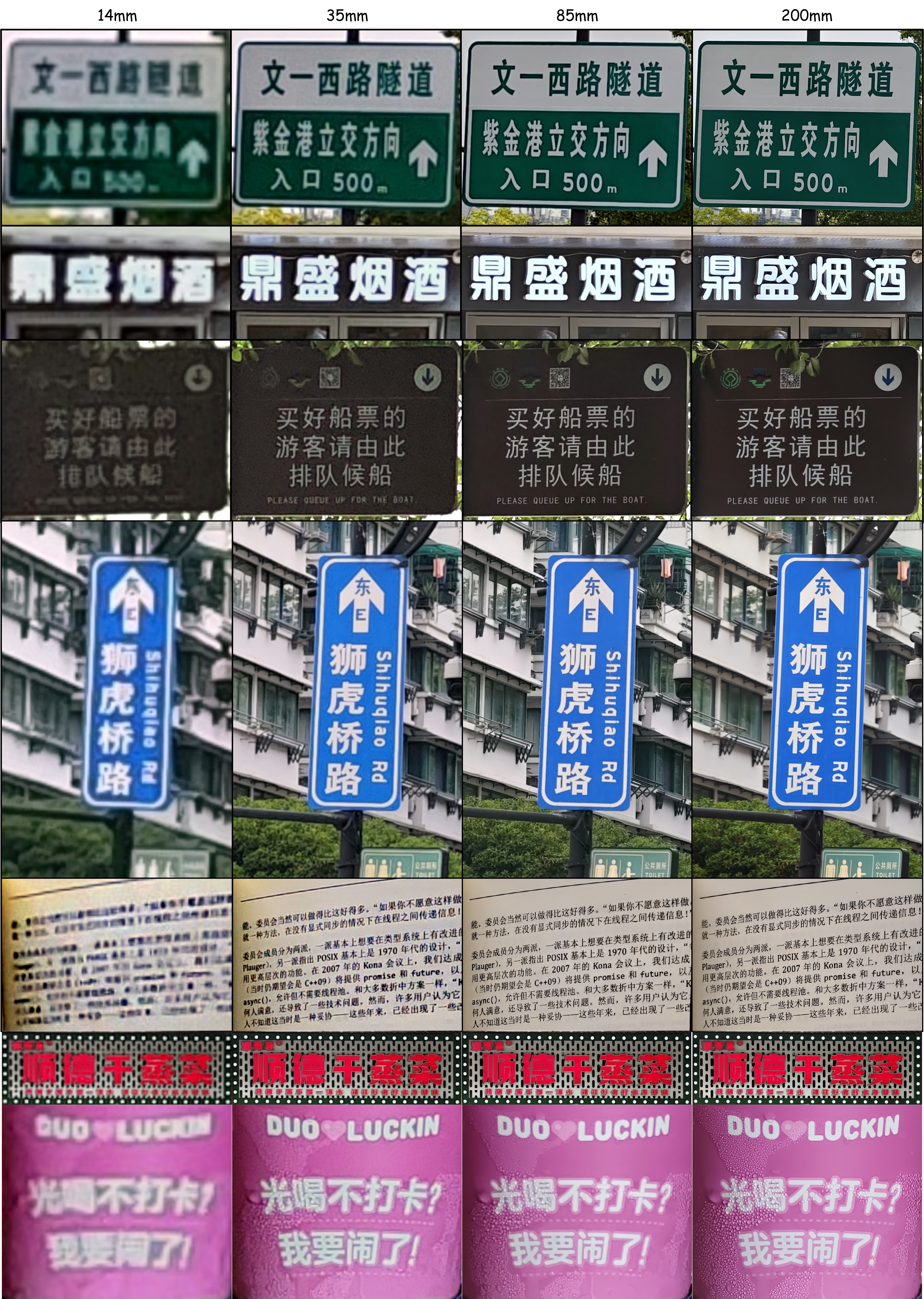}
\end{center}
\vspace{-10pt}
\caption{Detailed examples of UZ-ST.}

\label{fig:ultrazoom-details}
\end{figure*}

In \cref{tab:dataset_detail}, we provide detailed statistics on the composition of the UZ-ST dataset. 
Additionally, in \cref{fig:ultrazoom-details}, we present some example images from the dataset.

\begin{figure*}[t]
\begin{center}
\vspace{-5pt}
\includegraphics[width=1.0\linewidth]{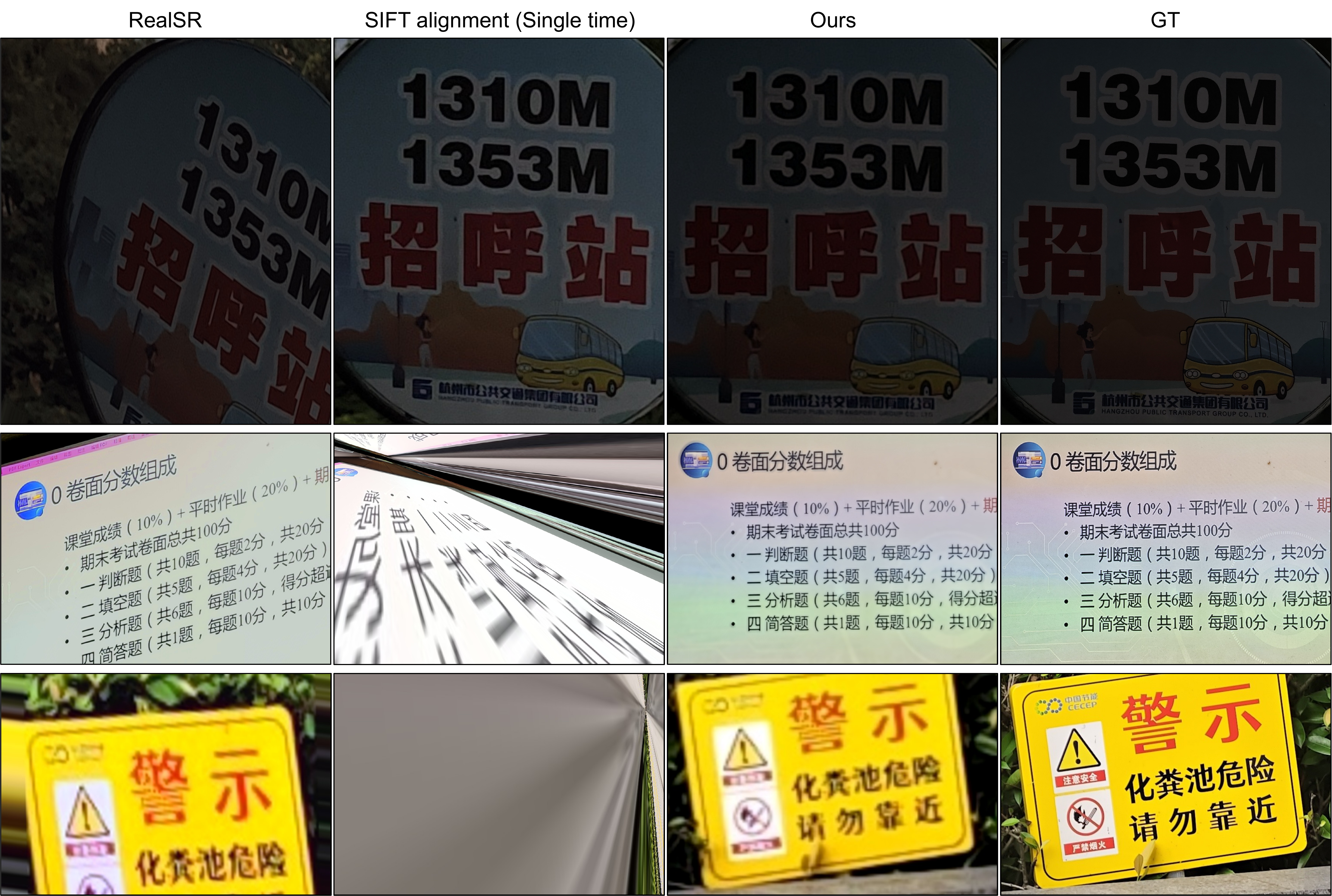}
\end{center}
\vspace{-10pt}
\caption{Alignment method comparison.}
\vspace{-10pt}
\label{fig:alignment}
\end{figure*}

\section{Alignment Comparison}

We compare our proposed Cascade Coarse-to-Fine alignment strategy with the alignment method proposed by RealSR~\cite{cai2019toward} and single-time alignment using SIFT.
During the comparison, we implement our strategy using SIFT and utilize the raw images from the 35mm dataset of our proposed UZ-ST dataset for evaluation.
For metrics, we adopt PSNR, Mean Squared Error (MSE), SSIM~\cite{wang2019textsr}, Normalized Cross-Correlation (NCC), and Average Keypoint Distance (AKD) to assess the alignment quality of the aligned image.

As shown in \cref{tab:align} and \cref{fig:alignment}, pixel-based optimization methods break when handling heavily degraded images due to significant loss of details. Keypoint-based alignment like SIFT suffers from the lack of valid keypoint for matching. 
While our alignment method outperforms all other methods and achieves the best results, this is due to the progressive cascade alignment strategy, which breaks the hard alignment process into simpler tasks, facilitating the bridging between images taken under different focal lengths.

\section{Additional results}
\label{sec:Appendix:additional}

\begin{figure*}[t]
\begin{center}
\vspace{-5pt}
\includegraphics[width=1.0\linewidth]{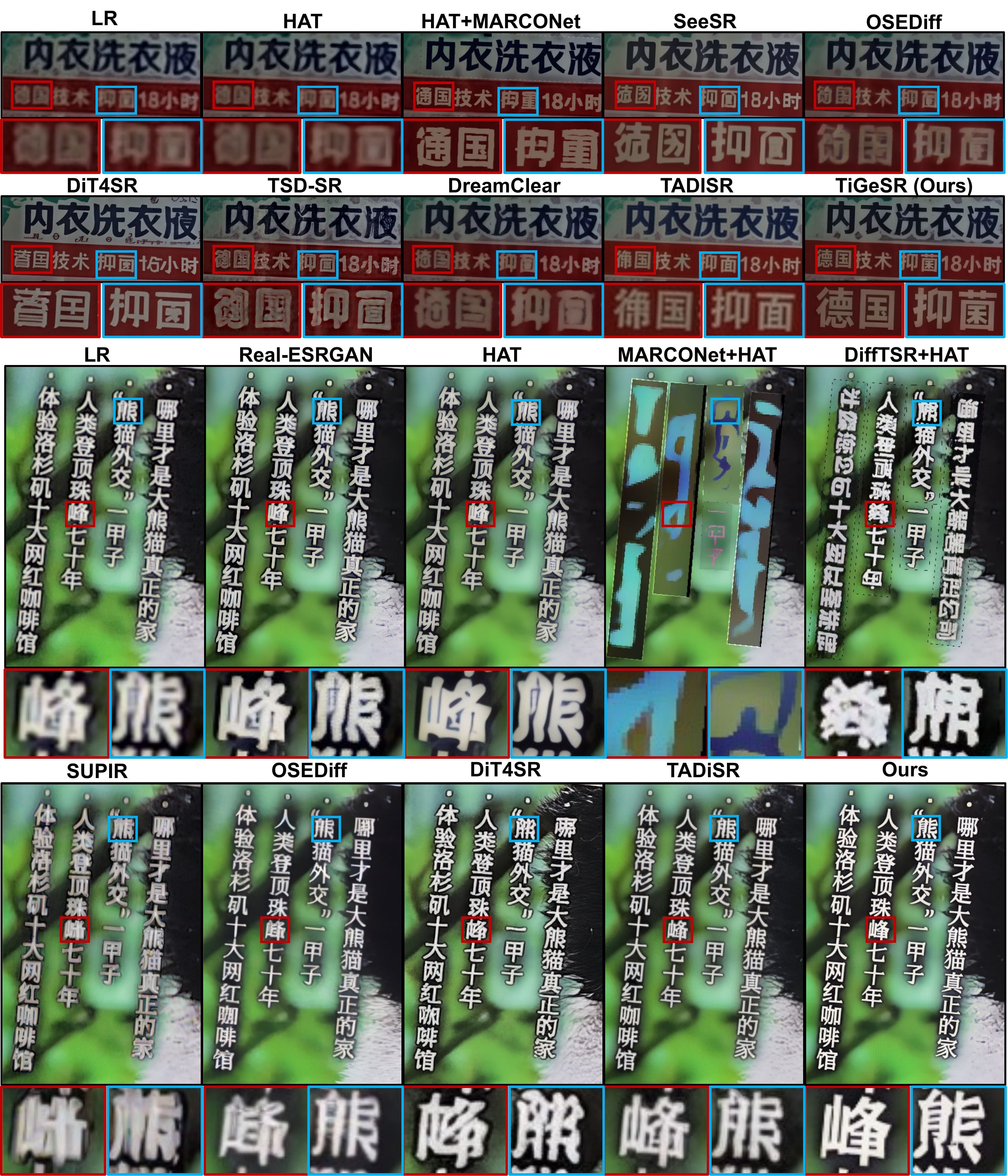}
\end{center}
\vspace{-10pt}
\caption{Additional qualitative results of TiGeSR.}
\vspace{-10pt}
\label{fig:Extraqua1}
\end{figure*}

\begin{figure*}[t]
\begin{center}
\vspace{-5pt}
\includegraphics[width=0.9\linewidth]{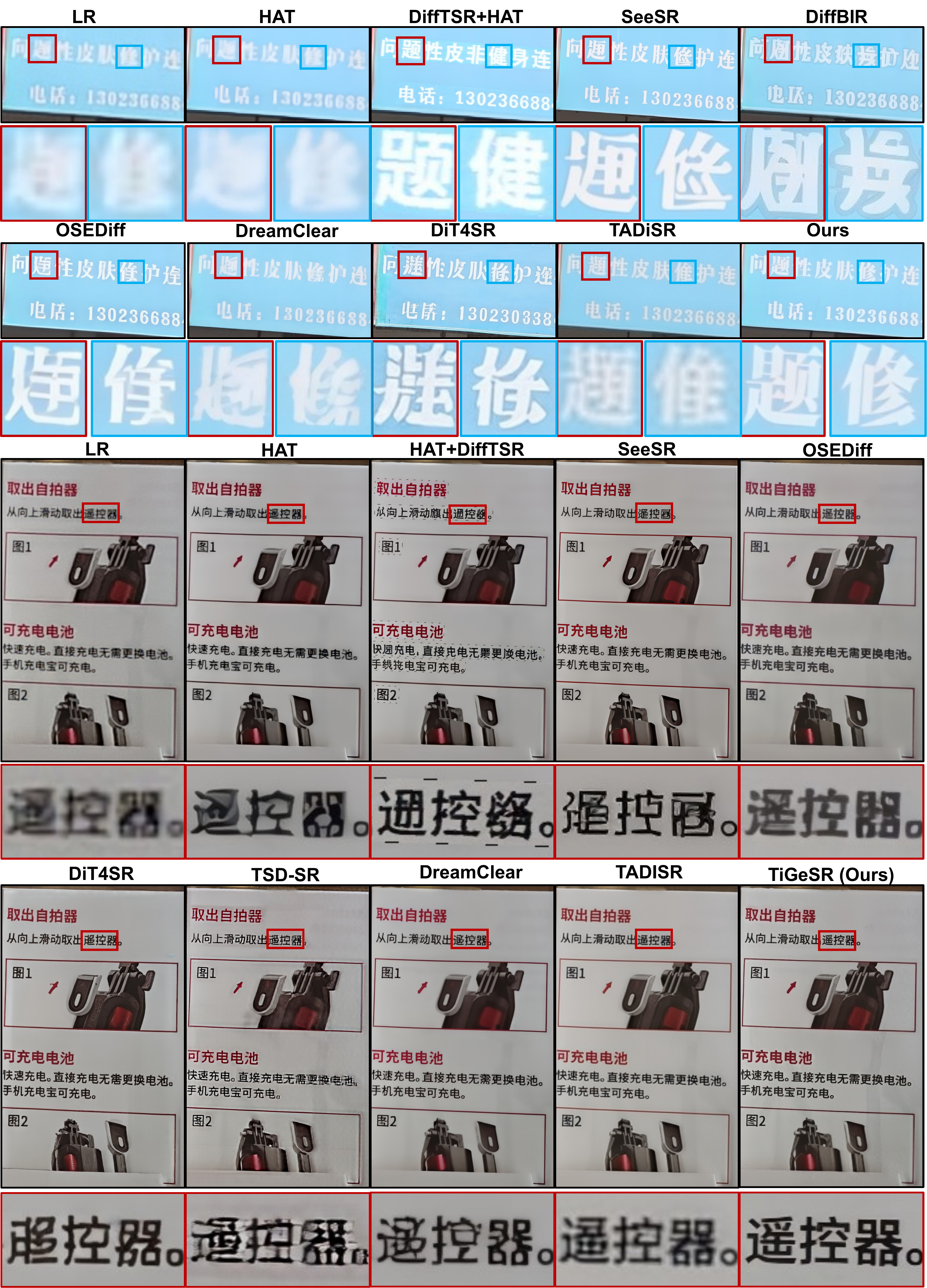}
\end{center}
\vspace{-10pt}
\caption{Additional qualitative results of TiGeSR.}
\vspace{-10pt}
\label{fig:Extraqua2}
\end{figure*}

\begin{figure*}[t]
\begin{center}
\vspace{-5pt}
\includegraphics[width=0.9\linewidth]{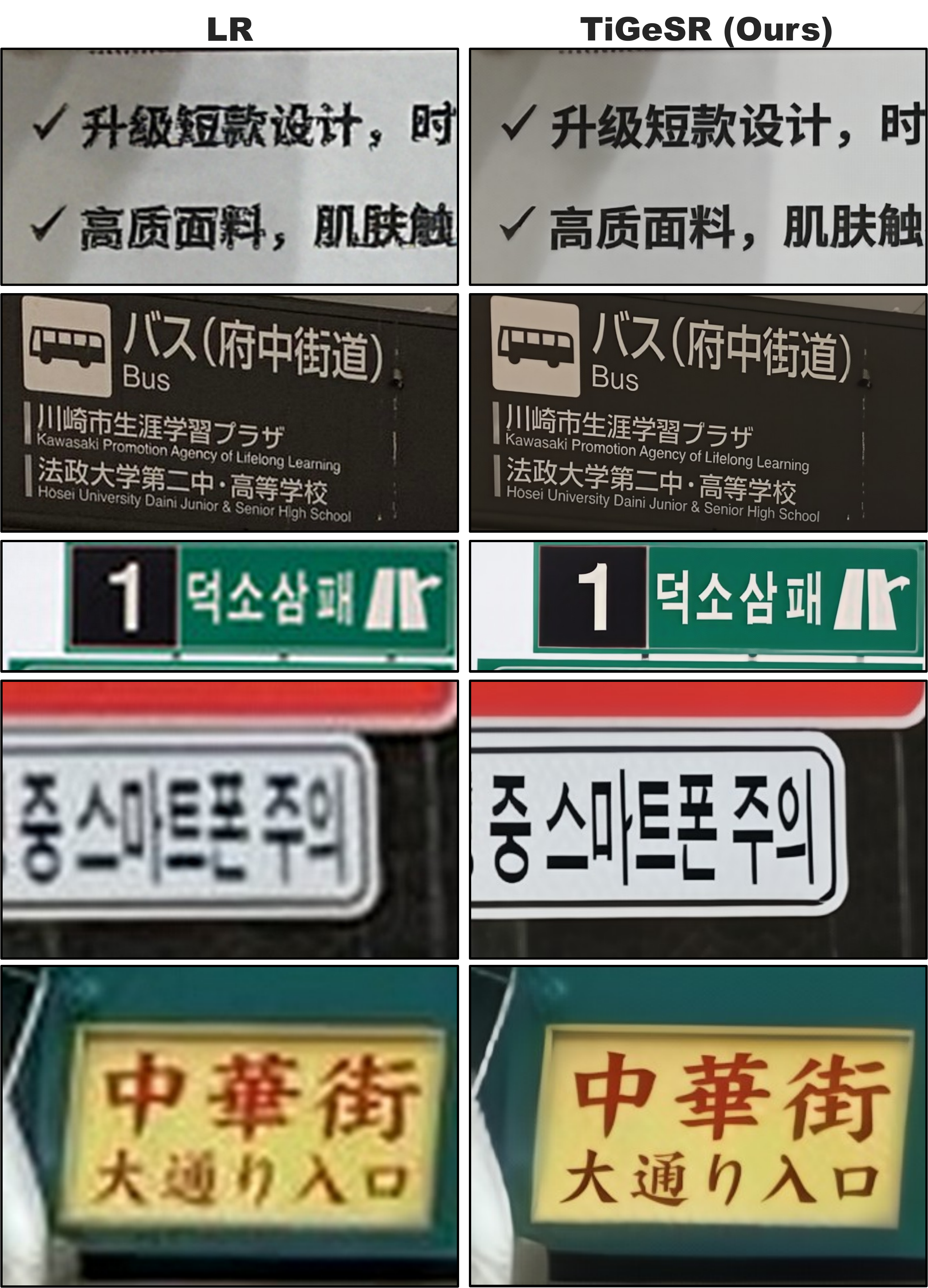}
\end{center}
\vspace{-10pt}
\caption{Demonstration of our method under different languages. (Zoom in for more details.)}
\vspace{-10pt}
\label{fig:multilingual}
\end{figure*}

We provide detailed quantitative results of UZ-ST in \cref{tab:Ultrazoom}. 
Additional qualitative results are provided in \cref{fig:Extraqua1} and \cref{fig:Extraqua2}.
\cref{fig:multilingual} also shows the ability of our method in handling other languages like Japanese and Korean. 

We also conduct additional ablation experiment on how the randomness of OCR output affect our method. 
As shown in \cref{tab:ocr_ablation_detail}, the text accuracy gains as the randomness of the OCR output declines. 
However, even under complete random OCR prediction, our method still manage to outperform TADiSR thanks to the LR condition still guiding the stage 1 to generate text structures.
This proves that the reliance of our method on the OCR model is limited.

\section{Efficiency Analysis}
As shown in \cref{tab:Efficiency}, stage 1 is a standard diffusion process that takes multiple steps in inference, the efficiency of our model may be suboptimal compared to one-step methods.
However, our method achieves state-of-the-art performance in OCR-A, which cannot be easily obtained by extending inference time.





\end{document}